%% file: paper-20260625-v2.tex
\documentclass{article}

\usepackage{PRIMEarxiv}
\usepackage[utf8]{inputenc}
\usepackage[T1]{fontenc}
\usepackage{hyperref}
\hypersetup{hidelinks,colorlinks=false,pdfborder={0 0 0}}
\usepackage{url}
\usepackage{booktabs}
\usepackage{amsfonts}
\usepackage{nicefrac}
\usepackage{microtype}
\usepackage{fancyhdr}
\usepackage{graphicx}
\usepackage{bm}
\usepackage{amssymb}
\usepackage{multirow}
\usepackage{amsmath}
\usepackage{pifont}
\usepackage{float}
\usepackage[table]{xcolor}
\usepackage{tabularx}
\usepackage{array}
\usepackage{tikz}
\usetikzlibrary{positioning,arrows.meta,fit,backgrounds,shapes.geometric}

\definecolor{trajectoryblue}{HTML}{2F80ED}
\definecolor{trajectorygreen}{HTML}{27AE60}

\newcommand{\cmark}{\ding{51}}
\newcommand{\xmark}{\ding{55}}
\title{CosFly-VLA: A Spatially Aware Vision-Language-Action Model for UAV Tracking}

\author{
  {\small\bfseries
  Ruilong Ren$^{1}$\footnotemark[1] \quad
  Songsheng Cheng$^{1}$\thanks{Equal contribution.} \quad
  Yunpeng Zhou$^{2}$ \quad
  Hanxuan Chen$^{1}$ \quad
  Xiangyue Wang$^{1}$\footnotemark[2]} \\[-0.15em]
  {\small\bfseries
  Tianle Zeng$^{3}$ \quad
  Shuai Yuan$^{4}$ \quad
  Binbo Li$^{4}$ \quad
  Hanzhong Guo$^{5}$ \quad
  Ji Pei$^{1}$ \quad
  Da Zhang$^{1}$\thanks{Corresponding authors.} \quad
  Kangli Wang$^{1}$\footnotemark[2]} \\[0.25em]
  {\normalfont\footnotesize
  $^{1}$Autel Robotics \quad
  $^{2}$Northeast Normal University \quad
  $^{3}$Southern University of Science and Technology} \\
  {\normalfont\footnotesize
  $^{4}$Peking University \quad
  $^{5}$University of Hong Kong} \\
  {\normalfont\footnotesize\texttt{wangxiangyue@autelrobotics.com}}
}

\begin{document}
\maketitle
\vspace{-1.0em}

\begin{abstract}
\input{chapters/abstract}
\end{abstract}


\begin{figure}[!t]
\centering
\includegraphics[width=\textwidth]{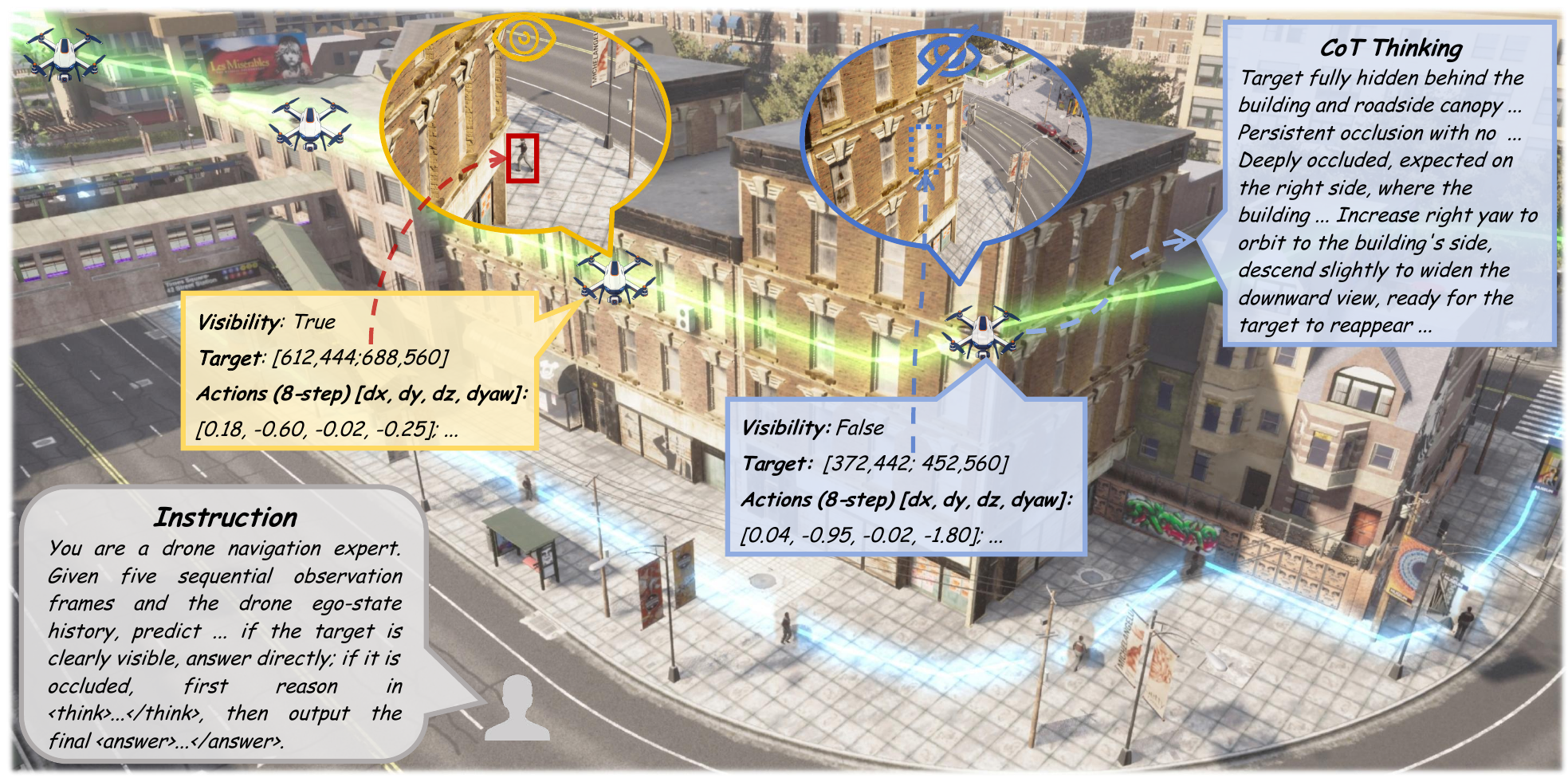}
\caption{
\textbf{Task overview.}
The \textcolor{trajectoryblue}{\textbf{blue trajectory}} denotes the pedestrian's motion, and the \textcolor{trajectorygreen}{\textbf{green trajectory}} denotes the UAV's executed flight path. At each step, the agent observes five egocentric frames, previous target boxes, and the task prompt. It predicts three synchronized outputs: target visibility, the current-frame bounding box, and an 8-step 4-DoF waypoint action chunk. When the target is visible, the policy can answer directly from visual evidence; when the target is occluded, CoT reasoning estimates the likely re-acquisition region before emitting the structured answer.
}
\label{fig:teaser}
\end{figure}

\input{chapters/01_introduction}
\input{chapters/02_related_work}
\input{chapters/03_task_formulation}
\input{chapters/04_method}

\input{chapters/05_experiments}
\input{chapters/06_discussion}
\input{chapters/07_limitations_and_futurework}
\input{chapters/08_conclusion}

\bibliographystyle{plain}
\bibliography{references}

\end{document}

%% file: chapters/abstract.tex
Dynamic target tracking is essential for Unmanned Aerial Vehicles (UAVs) operating in complex urban environments, where both the target and the camera viewpoint change continuously. Existing Vision-Language-Action (VLA) policies can track visible targets effectively, but their performance often degrades when buildings, vegetation, or roadside objects block the line of sight. During sustained occlusion, a policy may lose the target state, execute actions toward an incorrect region, and amplify this error through subsequent observations until re-acquisition becomes impossible. In this work, we formulate occlusion-robust UAV tracking as a closed-loop recovery problem: the policy must maintain a spatial hypothesis about an unseen target, predict where it may reappear, and correct errors induced by its own previous actions. 
To this end, we present \textbf{CosFly-VLA}, a spatially aware VLA model that jointly grounds the target, estimates its visibility, and generates continuous flight actions through a structured prediction interface. To train this policy, we use a large-scale recipe over diverse data sources. Spatially Grounded Continued Pretraining (CPT) on a 500k mixed pool injects UAV-view depth, distance, and 3-D spatial reasoning. A three-stage Curriculum-based Supervised Fine-Tuning (SFT) process then specializes the tracker through multi-head warm-up followed by two-stage curriculum learning over natural and hard / long-occlusion data. Chain-of-Thought (CoT) training subsequently teaches recovery-oriented reasoning traces before structured answers. Finally, a closed-loop Reinforcement Learning (RL) stage optimizes tracking behavior with a multi-component reward covering stand-off tracking, grounding quality, collision avoidance, and task success. 
We evaluate CosFly-VLA with occlusion-graded open-loop windows and closed-loop CARLA rollouts on seen and unseen maps. Relative to OpenVLA, CosFly-VLA-0.8B reduces open-loop Average Displacement Error (ADE) by \textbf{34.1\%} on seen-test and \textbf{35.3\%} on unseen-test. Closed-loop optimization improves Success Rate (SR) by \textbf{29.8\%} (\textbf{+17} percentage points) and \textbf{2.5\%} (\textbf{+2} points), respectively. These results demonstrate progress from visible-frame imitation toward spatially grounded action-closed-loop control, evaluated under a shared oracle state history.

%% file: chapters/01_introduction.tex
\section{Introduction}

Dynamic target tracking is a representative closed-loop embodied task for autonomous Unmanned Aerial Vehicles (UAVs). While continuously perceiving and localizing a moving target, a UAV must cope with viewpoint changes, relative-position changes, collision risks induced by its own motion, and it must adjust its flight strategy in real time. This capability is valuable for search-and-rescue, inspection, traffic monitoring, cinematography, delivery, and pursuit~\cite{khawaja2025survey,jayaweera2020dynamic,10445025,tian2025uavs}. In low-altitude urban flight, however, tracking failures often arise not from ordinary visible-frame localization errors, but from recurring occlusion. A target may be temporarily hidden by buildings, trees, vehicles, or crowds. Once the visual cue disappears from the field of view, conventional detection--re-identification (ReID)--control pipelines~\cite{8642452,Li_2020_CVPR,sun2023uav,wu2025learning,li2023adaptive,xue2024handling,Xue_2025_CVPR,sun2025refdrone} may lose the stable identity evidence needed for recovery. At that point, tracking is no longer simply a matter of ``seeing and following'' the target; it becomes a closed-loop decision problem under occlusion. The agent must infer where an unseen target may have moved, keep acting without direct visual observation, and re-acquire the target before tracking errors continue to accumulate.

Vision-Language-Action (VLA) models offer a natural modeling route for this setting. By coupling the semantic understanding and spatial priors of pretrained Vision-Language Models (VLMs) with policy modules that directly generate actions, VLA models have shown strong potential in robot manipulation, general control, and embodied navigation~\cite{brohan2023rt1roboticstransformerrealworld,brohan2023rt2visionlanguageactionmodelstransfer,kim2024openvlaopensourcevisionlanguageactionmodel,octomodelteam2024octoopensourcegeneralistrobot,black2026pi0visionlanguageactionflowmodel,intelligence2025pi05visionlanguageactionmodelopenworld,intelligence2025pi06vlalearnsexperience,zhai2025ignitingvlmsembodiedspace,song2025hume,li2024cogactfoundationalvisionlanguageactionmodel,cheang2025gr3technicalreport,li2025grrlgoingdexterousprecise,han2024dualprocessvlaefficient}. This paradigm has also been extended to UAV mission generation, fuzzy-instruction navigation, onboard inference, drone racing, aerial manipulation, and UAV vision-language tasks~\cite{sun2026airvlavisionlanguageactionsystemsaerial,sun2026autoflyvisionlanguageactionmodeluav,wu2025vlaanefficientonboardvisionlanguageaction,serpiva2025racevlavlabasedracingdrone,wang2025uavflowcolosseorealworldbenchmark,lykov2025cognitivedronevlamodelevaluation,10974117,liu2025indooruavbenchmarkingvisionlanguageuav,11149880,huang2026navdreamervideomodelszeroshot,chen2026visionlanguageuavs,chen2026cosfly}. Recent aerial tracking work, including UAV-Track VLA~\cite{zhang2026uavtrackvla}, TrackVLA-style embodied trackers, and open-world drone active tracking~\cite{pmlr-v305-wang25f,liu2025trackvlaunleashingreasoningmemory,wu2025hierarchicalinstructionawareembodiedvisual,dionigi2024d,sun2025openworld}, further shows that VLA policies can support complex tracking and pursuit tasks. Yet, from the perspective of sustained occlusion, existing studies have not made the sequence of target disappearance, reasoning under occlusion, and target re-acquisition the central object of training and evaluation. When the target leaves the field of view for an extended period, the model must use spatial priors to maintain reasonable actions, suppress drift caused by its own previous actions, and stably re-acquire the target after occlusion.

Sustained occlusion exposes three limitations in existing training pipelines. First, the model needs stronger aerial spatial reasoning. After the target disappears from the image, its likely reappearance depends on depth relations, relative scale, height difference, motion direction, and occlusion ordering under the UAV viewpoint; yet generic vision-language pretraining does not necessarily provide sufficient supervision for these UAV-specific geometric cues. Aerial spatial and reasoning datasets already provide related signals, spanning fine-grained grounding, embodied 3-D spatial concepts, multi-UAV perception, aerial mathematical reasoning, high-resolution aerial Visual Question Answering (VQA), and aerial captioning~\cite{airspatial2025,open3dvqa2025,aircop2025,avimath2025,hrvqa2024,capera2025}. They are usually treated, however, as perception or question-answering benchmarks rather than as pretraining supervision for an action-producing tracker. Second, the training data must preserve enough long-occlusion samples. Many window-construction pipelines discard samples in which the target is invisible in most historical frames, but such samples provide the supervision needed to learn post-occlusion recovery behavior. Third, target re-acquisition must be optimized in closed loop. Once the model makes an incorrect prediction at one step, its action changes the subsequent observation distribution. Recovery performance therefore cannot rely on offline supervision alone; it must be trained and evaluated on the state distribution induced by the policy's own rollouts.

Building on these observations, we present \textbf{CosFly-VLA}, an occlusion-recovery-oriented and spatially aware UAV VLA tracking framework (Figure~\ref{fig:teaser}). The task interface asks the policy to infer visibility, localize the current target box, and output an 8-step Four-Degree-of-Freedom (4-DoF) waypoint-delta action chunk from five Red-Green-Blue (RGB) frames, language instructions, and state history. CosFly-VLA uses Qwen3.5~\cite{qwen2026qwen35} as the vision-language backbone, extracts hidden states related to action planning, target localization, and visibility estimation through meta-query spans, and decodes the three structured outputs with decoupled heads. To strengthen spatial reasoning when the target is invisible, we introduce Spatially Grounded Continued Pretraining (CPT) over aerial spatial-reasoning data. The tracker is then specialized with a three-stage Curriculum-based Supervised Fine-Tuning (SFT) process: multi-head warm-up followed by two-stage curriculum learning over natural and hard / long-occlusion data. Chain-of-Thought (CoT) training then teaches recovery-oriented reasoning, and closed-loop Reinforcement Learning (RL) trains the policy on the observation distribution induced by its own actions, using rewards tied to stand-off tracking, grounding quality, collision avoidance, and task success.

To keep the evaluation aligned with the core failure mode, we use occlusion-graded open-loop windows and closed-loop CARLA rollouts over seen-test and unseen-test splits. Relative to OpenVLA, CosFly-VLA-0.8B (SFT+CoT) reduces open-loop Average Displacement Error (ADE) by 34.1\% / 35.3\% on the two splits; the cumulative training-recipe ablation associates the strongest cross-map gain with the step that adds spatially grounded CPT and the largest seen-test Hard gain with the step that adds CoT supervision. Also relative to OpenVLA, the RL-tuned variant improves closed-loop Success Rate (SR) by 29.8\% / 2.5\% and reduces rollout ADE by 10.0\% / 8.7\% on seen-test / unseen-test. 

The contributions of this paper are summarized as follows:
\begin{itemize}\itemsep=2pt
\item We revisit dynamic UAV target tracking from the perspective of sustained occlusion rather than visible-frame following. We formulate occlusion-robust tracking as a closed-loop recovery problem that requires target-state preservation, drift suppression, and post-occlusion re-acquisition, and define reproducible Easy / Medium / Hard protocols at both the open-loop window and closed-loop episode levels.

\item We introduce \textbf{CosFly-VLA}, a spatially aware VLA model for UAV target tracking. A meta-query interface connects shared multimodal representations to decoupled action, target-box, and visibility pathways, enabling the model to jointly perform continuous flight control, spatial grounding, and target-state estimation under partial observability.

\item We design a progressive training recipe that combines Spatially Grounded Continued Pretraining, three-stage Curriculum-based Supervised Fine-Tuning, dedicated Chain-of-Thought training, and closed-loop Reinforcement Learning. These stages successively introduce aerial geometric priors, structured tracking behavior, recovery-oriented reasoning, and adaptation to the observation distribution induced by the policy's own actions.

\item We evaluate CosFly-VLA on occlusion-graded seen-test and unseen-test splits through structured open-loop prediction and executed CARLA rollouts. Comparisons with general VLM/VLA baselines, cumulative training-recipe ablations, and qualitative trajectories demonstrate improvements in waypoint prediction, target grounding, closed-loop success, and cross-map generalization.
\end{itemize}

%% file: chapters/02_related_work.tex
\section{Related Work}
\label{sec:related}

\paragraph{VLA models for embodied AI.}
General-purpose VLA models align visual observations, language instructions, and action outputs within a unified policy, providing an important architectural foundation for CosFly-VLA. The RT series~\cite{brohan2023rt1roboticstransformerrealworld,brohan2023rt2visionlanguageactionmodelstransfer}, OpenVLA and Octo~\cite{kim2024openvlaopensourcevisionlanguageactionmodel,octomodelteam2024octoopensourcegeneralistrobot}, and the $\pi_0$ / $\pi_{0.5}$ / $\pi_{0.6}$ family~\cite{black2026pi0visionlanguageactionflowmodel,intelligence2025pi05visionlanguageactionmodelopenworld,intelligence2025pi06vlalearnsexperience} show how pretrained multimodal backbones can be coupled with robot-control modules to generate executable actions from visual and language inputs. Other studies add reasoning, memory, dual-process modeling, dexterous manipulation, action chunking, continuous action modeling, and flow-matching policies~\cite{han2024dualprocessvlaefficient,song2025hume,li2024cogactfoundationalvisionlanguageactionmodel,cheang2025gr3technicalreport,li2025grrlgoingdexterousprecise,zhai2025ignitingvlmsembodiedspace,zhao2023learningfinegrainedbimanualmanipulation}, while navigation and autonomous-driving systems adapt similar ideas to longer-horizon decisions~\cite{zhang2025uninavidvideobasedvisionlanguageactionmodel,bajcsy2024learning,alpamayo2025,jiang2025survey}. CosFly-VLA builds on this direction through a frozen vision-language backbone, a continuous flow-matching action expert, and task-specific hidden-state routing. Its focus, however, is different from general manipulation or navigation: the policy must continue acting while the target is invisible, and it must jointly predict motion, target location, and visibility for closed-loop recovery.

\paragraph{Embodied visual tracking and UAV active tracking.}
Visual tracking is increasingly viewed as a closed-loop problem that requires perception, decision-making, and control to be coordinated over time~\cite{Xue_2025_CVPR,xue2024handling,zhong2023rspt,9521193}. TrackVLA / TrackVLA++~\cite{pmlr-v305-wang25f,liu2025trackvlaunleashingreasoningmemory}, EZREAL~\cite{zeng2025ezreal}, HIET~\cite{wu2025hierarchicalinstructionawareembodiedvisual}, and related vision-language-model advisor and active-perception systems~\cite{11246600,Wu2025,Zhong_2025_ICCV,bajcsy2024learning} introduce language-conditioned policies, reasoning, and memory mechanisms into embodied tracking. These works mainly target ground robots or indoor agents, whose motion patterns, camera geometry, and occlusion structures differ from those of low-altitude UAVs. UAV tracking, in contrast, has often been organized around detection, redetection, re-identification, tracking, and control modules~\cite{8642452,Li_2020_CVPR,sun2023uav,wu2025learning,li2023adaptive,Xue_2025_CVPR,xue2024handling,sun2025refdrone}, with end-to-end active-tracking systems such as D-VAT~\cite{dionigi2024d} and GC-VAT~\cite{sun2025openworld} exploring learned pursuit policies. UAV-Track VLA~\cite{zhang2026uavtrackvla} further demonstrates that VLA modeling is promising for aerial active tracking. CosFly-VLA is complementary to these efforts: building on CosFly-Track~\cite{wang2026cosfly,chen2026track}, it makes sustained target disappearance and post-occlusion re-acquisition the organizing principle for data mining, evaluation, and closed-loop training. The model explicitly keeps long-occlusion windows, decouples action, bounding-box, and visibility prediction through meta-query routing, and optimizes recovery on the observation distribution induced by its own actions.

\paragraph{UAV vision-language navigation and aerial spatial data.}
Beyond active tracking, UAV VLA research has advanced in mission generation, fuzzy-instruction navigation, onboard inference, drone racing, and aerial manipulation~\cite{sun2026airvlavisionlanguageactionsystemsaerial,sun2026autoflyvisionlanguageactionmodeluav,wu2025vlaanefficientonboardvisionlanguageaction,serpiva2025racevlavlabasedracingdrone,wang2025uavflowcolosseorealworldbenchmark,lykov2025cognitivedronevlamodelevaluation,10974117,liu2025indooruavbenchmarkingvisionlanguageuav,11149880,huang2026navdreamervideomodelszeroshot,zeng2026carla,zeng2026can}. Related surveys and roadmaps situate these tasks within the broader framework of UAV autonomy~\cite{khawaja2025survey,jayaweera2020dynamic,tian2025uavs,10445025,jiang2025survey,chen2026visionlanguageuavs}. Systems such as AerialVLA~\cite{xu2026aerialvla}, WorldVLN~\cite{zhao2026worldvln}, AwareVLN~\cite{guo2026awarevln}, and UAV-Flow~\cite{wang2025uavflowcolosseorealworldbenchmark} mainly focus on instruction following, navigation decisions, or action learning toward static goals. They provide useful foundations for UAV vision-language control, but they are not centered on dynamic targets that may disappear for extended periods. At the same time, aerial spatial vision-language datasets provide supervision signals that are valuable for occlusion recovery: AirSpatial~\cite{airspatial2025} covers aerial grounding and metric relations, Open3DVQA-v2~\cite{open3dvqa2025} studies embodied 3-D spatial concepts, AirCopBench~\cite{aircop2025} targets multi-UAV perception, AVI-Math~\cite{avimath2025} emphasizes aerial mathematical reasoning, HRVQA~\cite{hrvqa2024} covers high-resolution aerial visual question answering, and CapERA~\cite{capera2025} provides aerial event captions. CosFly-VLA repurposes these sources from perception or question-answering benchmarks into continued-pretraining supervision for a closed-loop tracking policy, giving the backbone aerial-viewpoint priors over spatial relations, scale, and distance before supervised learning and reinforcement learning.

%% file: chapters/03_task_formulation.tex
\section{Task Formulation}
\label{sec:task}

\paragraph{Task definition.}
We formulate UAV dynamic target tracking as the closed-loop pursuit of a moving target by a single quadrotor agent equipped with an egocentric monocular camera. The task is intentionally structured as a joint perception-and-control problem. At decision step $t$, the policy $\pi_\theta$ receives a short visual history, a language task / target prompt, and textual state history, then emits the current target grounding, an action chunk, and a visibility estimate, as summarized in Eq.~\ref{eq:task_io}:
\begin{equation}
\label{eq:task_io}
o_t=(I_{t-K+1{:}t},\, l,\, s_t^{\text{prev}}),\qquad
\hat y_t=(\hat b_t,\,\Delta_{t+1{:}t+H_a},\,\hat v_t).
\end{equation}

Here $I_{t-K+1{:}t}$ is a $K{=}5$-frame RGB stream at $640{\times}360$ resolution (width $\times$ height, with $H{=}360$ and $W{=}640$), $l$ is the language task / target prompt, and $s_t^{\text{prev}}$ contains 4-DoF UAV poses, previous target boxes, and visibility flags. The output $\hat y_t$ consists of the current Qwen-style target box $\hat b_t$, an $H_a{=}8$ future waypoint-delta action chunk $\Delta_{t+1{:}t+H_a}\in\mathbb{R}^{H_a\times4}$, and a target-visibility probability $\hat v_t$. Each action delta is $(\Delta x,\Delta y,\Delta z,\Delta\mathrm{yaw})$ in the controller's world-frame convention. At deployment, we use a receding-horizon controller with $H_c{=}1$: only the first predicted delta is executed before the next observation arrives. Unless otherwise stated, $s_t^{\text{prev}}$ is populated from ground-truth state history in both open-loop and closed-loop evaluation; robustness to predicted-state feedback is left to future work (Section~\ref{sec:limitations}).

\paragraph{Closed-loop occlusion-graded difficulty.}
Because occlusion is the central problem we target, the closed-loop protocol grades each episode by a geometric visibility test rather than by model predictions. At frame $t$, we cast five line-of-sight (LOS) rays from the UAV camera position to the target center, head, feet, left side, and right side. A ray is counted as blocked if scene geometry intersects it at least $0.5$\,m before the target point. Let $\mathrm{vis\_frac}_t$ be the fraction of the five rays that remain unblocked, and let $d_t$ be the UAV--target distance in meters. A frame is marked invisible if fewer than three rays are unblocked or the target falls outside the valid distance gate $(2,50)$\,m. For an episode of $T$ frames, define
\begin{equation}
\label{eq:occ_rate}
\chi_t^{\mathrm{occ}}=\mathbf{1}\!\left[\mathrm{vis\_frac}_t<0.6\ \lor\ d_t\notin(2,50)\right],
\qquad
\mathrm{occ\_rate}=\frac{1}{T}\sum_{t=1}^{T}\chi_t^{\mathrm{occ}}.
\end{equation}
We use \textbf{Easy} for $\mathrm{occ\_rate}<5\%$, \textbf{Medium} for $5\%\le\mathrm{occ\_rate}<20\%$, and \textbf{Hard} for $\mathrm{occ\_rate}\ge20\%$; Hard episodes with $\mathrm{occ\_rate}>70\%$ are filtered because nearly always-invisible targets do not define a meaningful tracking problem. We also record the longest contiguous occlusion duration, $\mathrm{occ\_streak\_s}$, as an auxiliary diagnostic. This episode-level protocol is distinct from the open-loop window-level grading in Section~\ref{sec:setup}, but both are designed to separate ordinary visible-target following from the recovery regimes that motivate CosFly-VLA.

%% file: chapters/04_method.tex
\section{Method}
\label{sec:method}

\subsection{Framework Overview}

\begin{figure}[t]
\centering
\includegraphics[width=\linewidth]{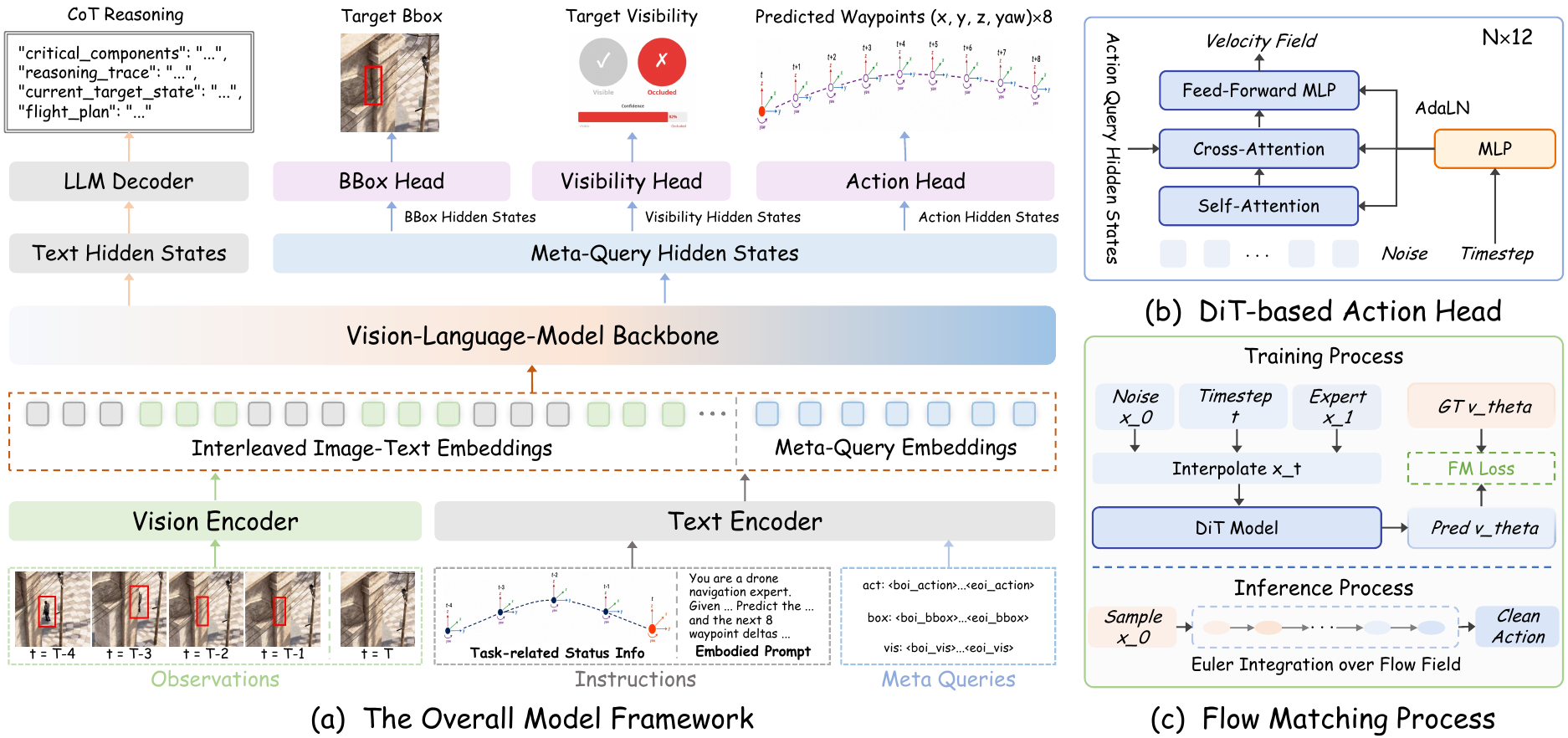}
\caption{
\textbf{CosFly-VLA architecture.}
Multi-frame observations, language instructions, state history, and boundary-tagged meta-query tokens are encoded by a frozen Qwen3.5 backbone with LoRA adapters. The meta-query hidden states are routed to three structured prediction heads: a 12-layer flow-matching DiT action expert for the 8-step 4-DoF waypoint-delta action chunk, a MLP bounding-box head, and a visibility head. The reasoning path shown in the diagram corresponds to CoT text generation and is not used as an additional control head. The right panels show the DiT action head and its flow-matching training / inference process.
}
\label{fig:arch}
\end{figure}

CosFly-VLA converts the closed-loop tracking state in Section~\ref{sec:task} into three synchronized outputs: the current target box, an 8-step 4-DoF waypoint-delta action chunk, and the target-visibility score. This decomposition matches the structure of UAV tracking: grounding anchors the visible target, action prediction drives the controller, and visibility estimation tells the policy when recovery should rely more on spatial and temporal context than on appearance.
The framework keeps Qwen3.5 as a frozen backbone $\Phi$ with Low-Rank Adaptation (LoRA)~\cite{hu2022lora}, and attaches three lightweight task heads through a meta-query interface (Figure~\ref{fig:arch}). The backbone encodes multi-frame RGB observations, the language prompt, UAV pose history, and previous boxes into a shared visual-text-state representation; meta-query spans then route task-specific hidden states to a flow-matching Diffusion Transformer (DiT) action expert, a Multilayer Perceptron (MLP) bounding-box (bbox) head, and an MLP visibility head. This avoids brittle JSON-like text decoding by producing structured boxes, visibility scores, and waypoint chunks directly from task-specific hidden states.

\subsection{Model Architecture}
\label{sec:meta_query}

CosFly-VLA adopts a backbone--heads architecture that separates shared multimodal understanding from task-specific continuous prediction. The visual-language backbone is responsible for encoding the image history, language prompt, UAV pose history, and previous target boxes into a unified representation. The task-specific heads then read different subsets of this representation through meta-query spans and decode the quantities required by closed-loop tracking: the 8-step waypoint-delta action chunk, the current target box, and target visibility. This organization preserves the general reasoning capacity of Qwen3.5 while giving the action, grounding, and visibility pathways enough specialization to handle the different failure modes of UAV tracking.

\subsubsection{Visual and Language Backbone}

CosFly-VLA uses Qwen3.5 as the shared visual-language backbone for high-level scene understanding, target semantics, and state-history integration. Given the observation tuple $o_t=(I_{t-K+1{:}t}, l, s_t^{\mathrm{prev}})$, the model receives multi-frame RGB observations, a language task / target prompt, historical UAV poses, and previous target boxes. These heterogeneous inputs are organized as a unified multimodal sequence, so the backbone can jointly encode visual appearance, linguistic target intent, and temporal state cues. This shared representation is essential for UAV tracking: when the target is visible, the model can rely on appearance and grounding; when the target is occluded, it must instead use scene layout, historical motion, and language-conditioned target identity to maintain a recoverable belief state.

We keep the Qwen3.5 backbone frozen across the task-specific stages (SFT, CoT, and RL) and adapt it with LoRA adapters ($r{=}16$, $\alpha{=}32$, dropout $0.05$ on the query / key / value / output projections of every attention layer). This design preserves the backbone's pretrained multimodal reasoning capacity while limiting the number of trainable parameters introduced by the UAV tracking task. The remaining trainable interface is deliberately small: newly introduced meta-query embeddings and lightweight connectors translate the shared hidden sequence into task-specific representations. This separation avoids over-specializing the backbone to one control convention and makes the same representation usable by the action, grounding, and visibility branches.

\begin{figure*}[!t]
\centering
\includegraphics[width=\textwidth]{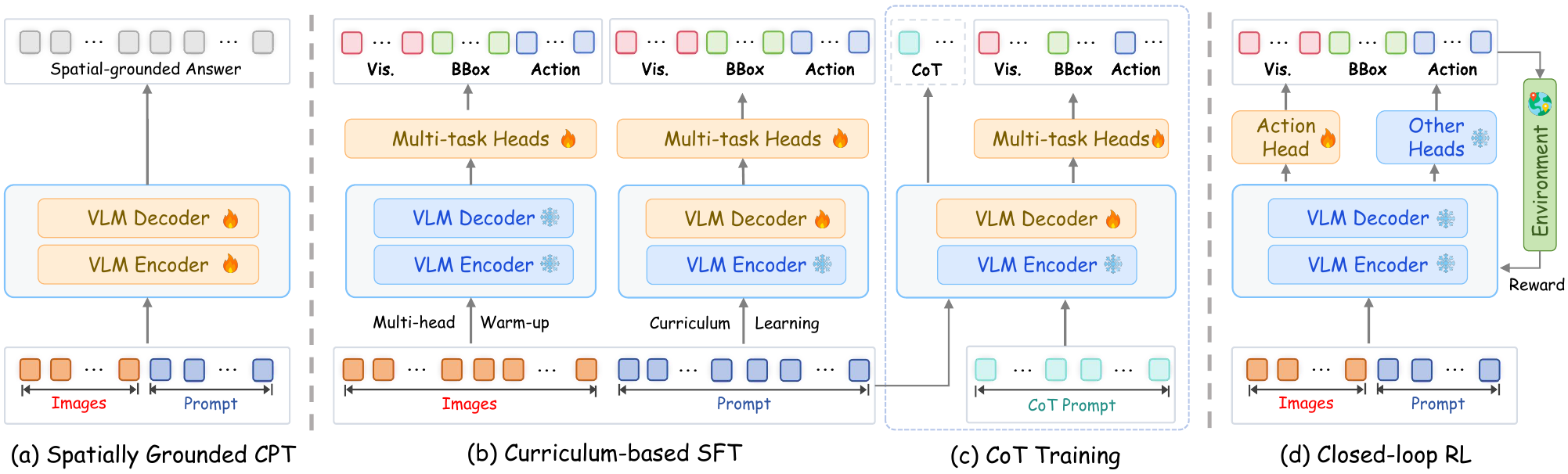}
\caption{
\textbf{CosFly-VLA training recipe.}
\emph{(a) Spatially Grounded CPT} adapts the Qwen3.5 backbone with aerial spatial-perception supervision.
\emph{(b) Curriculum-based SFT} is a three-stage supervised process: multi-head warm-up, natural-distribution curriculum learning, and hard / long-occlusion curriculum learning.
\emph{(c) CoT Training} teaches the model to emit recovery-oriented reasoning traces before structured answers.
\emph{(d) Closed-loop RL} updates the action head through interaction with the environment and tracking-oriented rewards, while the other components remain fixed.
}
\label{fig:training_pipeline}
\end{figure*}

\subsubsection{Task-Specific Heads}

\paragraph{Meta-query routing.}
To expose task-specific hidden states without serializing numeric answers through the language channel, we introduce three boundary-tagged meta-query groups, one per prediction family $\tau\in\{\text{act},\text{box},\text{vis}\}$, with span lengths $M_a{=}16$, $M_b{=}16$, and $M_v{=}1$:
\begin{equation}
\label{eq:meta_query_tokens}
[\texttt{<boi\_}\tau\texttt{>}]\;\;q^{\tau}_{1},\,q^{\tau}_{2},\,\dots,\,q^{\tau}_{M_\tau}\;\;[\texttt{<eoi\_}\tau\texttt{>}],
\qquad \tau\in\{\text{act},\text{box},\text{vis}\}.
\end{equation}
The query block is placed after the \texttt{<think>}\dots\texttt{</think>} block but before \texttt{<answer>}, so the task heads can condition on the reasoning trace without seeing the numeric answer tokens. Let $h\in\mathbb{R}^{B\times L\times d}$ denote the final backbone hidden states. We extract the task-specific spans
\begin{equation}
\label{eq:meta_query_extract}
h^{(\tau)} = \mathrm{Pick}(h, \langle\text{boi\_}\tau\rangle, \langle\text{eoi\_}\tau\rangle)\in\mathbb{R}^{B\times M_\tau\times d},
\quad \tau\in\{\text{act},\text{box},\text{vis}\},
\end{equation}
and pass each span through a connector $C^{(\tau)}$ with Root Mean Square Layer Normalization (RMSNorm) and a Linear--Gaussian Error Linear Unit (GELU)--Linear projection, yielding $z^{(\tau)}=C^{(\tau)}(h^{(\tau)})$. The evaluated 0.8B configuration uses a 2-layer / 8-head Transformer connector; scaling to larger backbones would use a shallower connector to fit the memory budget, which we leave to future work. The connector is the architectural boundary between shared Qwen3.5 features and task-specific decoding: action queries focus on future control, bbox queries focus on spatial grounding, and the single visibility query focuses on whether appearance evidence should be trusted.

\paragraph{Flow-matching action expert.}
\label{sec:action_head}
The action branch is responsible for producing the 8-step 4-DoF waypoint-delta action chunk consumed by the UAV controller. We implement $\Psi^{\text{act}}$ as a 12-layer DiT-style flow-matching expert with adaptive Layer Normalization (adaLN) modulation~\cite{peebles2023dit} and per-block cross-attention to $z^{\text{act}}$. Instead of directly regressing a deterministic displacement vector, the expert learns a rectified-flow velocity field over normalized action chunks. This is better suited to UAV tracking than text-token action decoding because it models a continuous trajectory distribution while remaining conditioned on the same language and visual context as the grounding heads. At inference, explicit Euler integration generates the action chunk, and a kinematic-feasibility clamp $\mathrm{KC}(\cdot)$ limits velocity, acceleration, and yaw-rate components before the first delta is executed by the receding-horizon controller.

\paragraph{Grounding and visibility heads.}
The bbox and visibility branches provide the visual grounding signals that close the loop between perception and action. They are intentionally lightweight because their role is to decode structured state from task-specific hidden representations rather than to perform another round of language modeling. The bbox head predicts the current target location needed for visual feedback, while the visibility head estimates whether the current observation contains reliable appearance evidence. The two heads operate on their connector outputs as
\begin{equation}
\label{eq:box_vis_heads}
\hat b_t = \Psi^{\text{box}}(\mathrm{mean}(z^{\text{box}})),\qquad
\hat v_t = \operatorname{sigmoid}(\Psi^{\text{vis}}(z^{\text{vis}}_1)).
\end{equation}
$\Psi^{\text{box}}$ and $\Psi^{\text{vis}}$ are 2-layer MLPs with GELU and LayerNorm. Box targets follow the Qwen grounding convention on a $0$--$1000$ grid and are normalized internally as $b/1000$ rather than raw pixels. The visibility head outputs a probability that is used both as a supervised signal and as a diagnostic of whether the controller should trust appearance-based evidence. This is particularly important in long-occlusion windows, where a high-quality policy must keep acting even when the current frame no longer contains a reliable target observation.

\subsection{Training Recipe}

Training CosFly-VLA requires more than fitting waypoint and bbox labels on visible-target frames. The policy must first acquire UAV-view spatial priors, then learn the structured tracking interface and recovery-oriented reasoning traces, and finally practice recovery under its own closed-loop feedback. Therefore, we use four successive phases (as shown in Figure~\ref{fig:training_pipeline}): spatially grounded CPT, three-stage curriculum SFT, CoT training, and closed-loop RL. CPT adapts the full backbone (including the vision encoder) to aerial depth, distance, and 3-D relation cues; SFT then freezes the backbone and specializes the decoupled heads through multi-head warm-up and two-stage curriculum learning; CoT training teaches reasoning traces before structured answers; and RL exposes the policy to drift, collision, and recovery events on-policy that do not appear in ordinary offline supervision.

\paragraph{Spatially grounded continued pretraining.}
\label{sec:cpt}
The first phase addresses a missing prerequisite for occlusion recovery: when the target is not visible, the policy must rely on scene geometry rather than appearance. Generic Qwen3.5 pretraining provides broad multimodal understanding, but it is not specifically optimized for UAV-view spatial relations, object scale, altitude cues, red-box grounding, or occlusion ordering. We therefore use spatially grounded continued pretraining as an upstream adaptation stage before any task-specific action head is trained.

The CPT data uses internal UAV trajectories as the main source because those samples are closest to dynamic tracking, red-box grounding, long-horizon reasoning, and hard negative recognition. It adds six external aerial spatial-VL datasets to broaden the backbone's aerial semantics and geometry priors. The final mixture is balanced to roughly 500k samples, with 70\% internal trajectory samples and 30\% external aerial data. All sources are converted into the same JavaScript Object Notation Lines (JSONL) schema, with fields such as images, messages, task, source, and extra metadata kept compatible with the downstream SFT entry point. CPT updates all backbone parameters, including the vision encoder; the resulting weights are saved directly as a spatially aligned backbone for the supervised curriculum.

CPT is designed to inject five capabilities that are difficult to learn from visible-frame tracking windows alone: aerial 3-D spatial relations, metric distance and size estimation, object attribute grounding, aerial mathematical / counting reasoning, and caption-level scene description. Table~\ref{tab:cpt_data} summarizes the data composition and the role of each source.
Table~\ref{tab:training_data_stats} summarizes the scale and role of each training phase, including the CoT reasoning data.

\begin{table}[!t]
\centering
\small
\setlength{\tabcolsep}{4pt}
\caption{\textbf{CPT-stage data composition.} Spatially grounded CPT uses a 70\% / 30\% mixture of internal UAV trajectory data and external aerial spatial-VL data. External datasets are used only for CPT, not for the SFT training pool.}
\label{tab:cpt_data}
\begin{tabularx}{\columnwidth}{@{}l c X@{}}
\toprule
\textbf{Source} & \textbf{Samples} & \textbf{Role in CPT} \\
\midrule
Internal UAV trajectory pipeline & 350{,}000 & UAV tracking, grounding, free-form captioning, negative examples, long-horizon look-ahead \\
AirSpatial~\cite{airspatial2025} & 45{,}000 & Aerial grounding, distance, size, color, type and 3-D relation cues \\
Open3DVQA-v2~\cite{open3dvqa2025} & 37{,}500 & Embodied 3-D spatial concepts and perspective reasoning \\
HRVQA~\cite{hrvqa2024} & 30{,}000 & High-resolution remote-sensing QA and scene reasoning \\
AVI-Math~\cite{avimath2025} & 15{,}000 & Aerial geometry, counting and algebra reasoning \\
AirCopBench~\cite{aircop2025} & 11{,}250 & Multi-UAV collaborative perception and reasoning \\
CapERA~\cite{capera2025} & 11{,}250 & Aerial single-frame description / captioning \\
\midrule
\textbf{Total} & $\sim$500{,}000 & Balanced by task and answer bucket \\
\bottomrule
\end{tabularx}
\end{table}

\begin{table}[t]
\centering
\small
\setlength{\tabcolsep}{4pt}
\caption{\textbf{Training data statistics.} The release follows CPT $\rightarrow$ three-stage curriculum SFT $\rightarrow$ CoT training $\rightarrow$ closed-loop RL. SFT is Town10HD pedestrian-only in this baseline release; RL uses a separate multi-map split.}
\label{tab:training_data_stats}
\begin{tabularx}{\columnwidth}{@{}l c X@{}}
\toprule
\textbf{Phase} & \textbf{Scale} & \textbf{Data / split role} \\
\midrule
CPT & $\sim$500k samples & 70\% internal UAV trajectory data + 30\% six external aerial spatial-VL datasets \\
Three-stage SFT & 5{,}861 clean trajectories / 1{,}740{,}278 samples & Town10HD pedestrian-only pool; multi-head warm-up + natural curriculum + hard / long-occlusion curriculum \\
Long-occ mining & 233{,}437 candidates & Relaxed miner admits 1{,}307 windows with 4-of-5 invisible inputs and 5{,}254 with 5-of-5 invisible inputs \\
CoT training & 19{,}066 samples & Teacher-generated reasoning traces for hard / long-occlusion recovery; $\lambda_{\text{lm}}=0.1$ \\
RL & 550 paths + 20 calibration / 104{,}250 frames & 11-map split: 440 train + 110 test; 20 historical calibration paths are separate \\
\bottomrule
\end{tabularx}
\end{table}

\paragraph{Curriculum-based supervised fine-tuning.}
\label{sec:curriculum}
The second phase turns the spatially aligned backbone into a structured UAV tracking policy. A single end-to-end SFT run is unstable because the action, bbox, and visibility paths have different loss scales, convergence speeds, and sample requirements. We therefore use a three-stage curriculum. First, \emph{multi-head warm-up} initializes the action, bbox, and visibility paths with the backbone frozen and LoRA disabled, so the task heads learn to read useful hidden states before backbone adaptation begins. Second, the warmed heads are reloaded into a unified multitask model, LoRA is enabled, and the policy is trained on the natural tracking distribution. Third, the curriculum shifts the data distribution toward rare maneuvers and occlusion-heavy windows, including an explicit 15\% long-occlusion quota. This three-stage SFT process trains the structured tracker; CoT training then follows as a dedicated reasoning phase rather than as an additional SFT stage.

Buckets in Stage 3 are assigned by a fixed priority rule over the $K{+}H_a$ window:
\begin{equation}
\label{eq:bucket}
\begin{aligned}
g(\cdot) = \mathrm{argmax}\,\bigl[\,&\text{long\_occ}\succ\text{avoid\_tight}\succ\text{avoid\_loose}\succ\text{sharp\_turn} \\
                                  \succ\,&\text{short\_occ}\succ\text{turn}\succ\text{small/alt}\succ\text{cruise}\,\bigr].
\end{aligned}
\end{equation}
The key indicators are \texttt{long\_occ} ($\geq 4$ invisible frames in the 5-frame window), \texttt{sharp\_turn} (cumulative yaw $\geq 30^\circ$), \texttt{avoid\_tight} (small obstacle distance while heading changes), and altitude / small-target cues. Stage-3 weights follow a $1{:}1.5{:}2{:}3$ ratio over cruise / turn / loose-avoidance / tight-avoidance buckets, so samples that most often lead to closed-loop failure exert larger training pressure than ordinary cruising windows.

The supervised objective follows the same multi-head structure. The action head is trained with a rectified-flow target over normalized action chunks. Given a ground-truth waypoint tensor $x_1\in[-1,1]^{H_a\times 4}$ with $H_a{=}8$, Gaussian noise $x_0\sim\mathcal{N}(0,I)$, and $t\sim\mathcal{U}[0,1]$, we form $x_t=(1-t)x_0+tx_1$ and train the DiT to predict the constant flow direction:
\begin{equation}
\mathcal{L}_{\text{act}} = \mathbb{E}_{x_1,x_0,t}\left[\bigl\|\Psi^{\text{act}}(x_t,t,z^{\text{act}}_{\text{pool}},z^{\text{act}})-(x_1-x_0)\bigr\|_2^2\right],
\label{eq:flow}
\end{equation}
where $z^{\text{act}}_{\text{pool}}=\mathrm{mean}(z^{\text{act}})$ conditions the global adaLN path and the full $z^{\text{act}}$ sequence supplies cross-attention keys and values. The bbox head uses an $\ell_1$ term plus Generalized Intersection-over-Union (GIoU) on normalized Qwen-style coordinates, and the visibility head uses binary cross-entropy. The full SFT loss is
\begin{equation}
\mathcal{L} = \sum_{i\in\mathcal{B}} w(g_i)\bigl[\lambda_a\mathcal{L}^{(i)}_{\text{act}} + \lambda_b\mathcal{L}^{(i)}_{\text{box}} + \lambda_v\mathcal{L}^{(i)}_{\text{vis}}\bigr] + \lambda_{\text{lm}}\mathcal{L}_{\text{lm}},
\label{eq:loss}
\end{equation}
where $w(g_i)$ is the curriculum bucket weight. Unless otherwise stated during SFT, $(\lambda_a,\lambda_b,\lambda_v,\lambda_{\text{lm}})=(1.0,1.0,0.05,0.0)$. The CoT phase sets $\lambda_{\text{lm}}=0.1$, with the language-modeling loss masked away from meta-query positions, so reasoning supervision can be learned without routing control supervision via the language head.

\paragraph{Chain-of-thought training.}
\label{sec:cot_training}
After the structured SFT curriculum, we train on CoT samples generated by a Qwen3.5-397B teacher~\cite{qwen2026qwen35} in 8-bit floating-point (FP8) precision. These samples ask the assistant to produce a reasoning trace in \texttt{<think>}\dots\texttt{</think>} before the structured \texttt{<answer>}. This phase teaches interpretable recovery reasoning for hard and long-occlusion cases while keeping action, bbox, and visibility supervision attached to the decoupled heads.

\begin{figure}[t]
\centering
\includegraphics[width=0.8\linewidth]{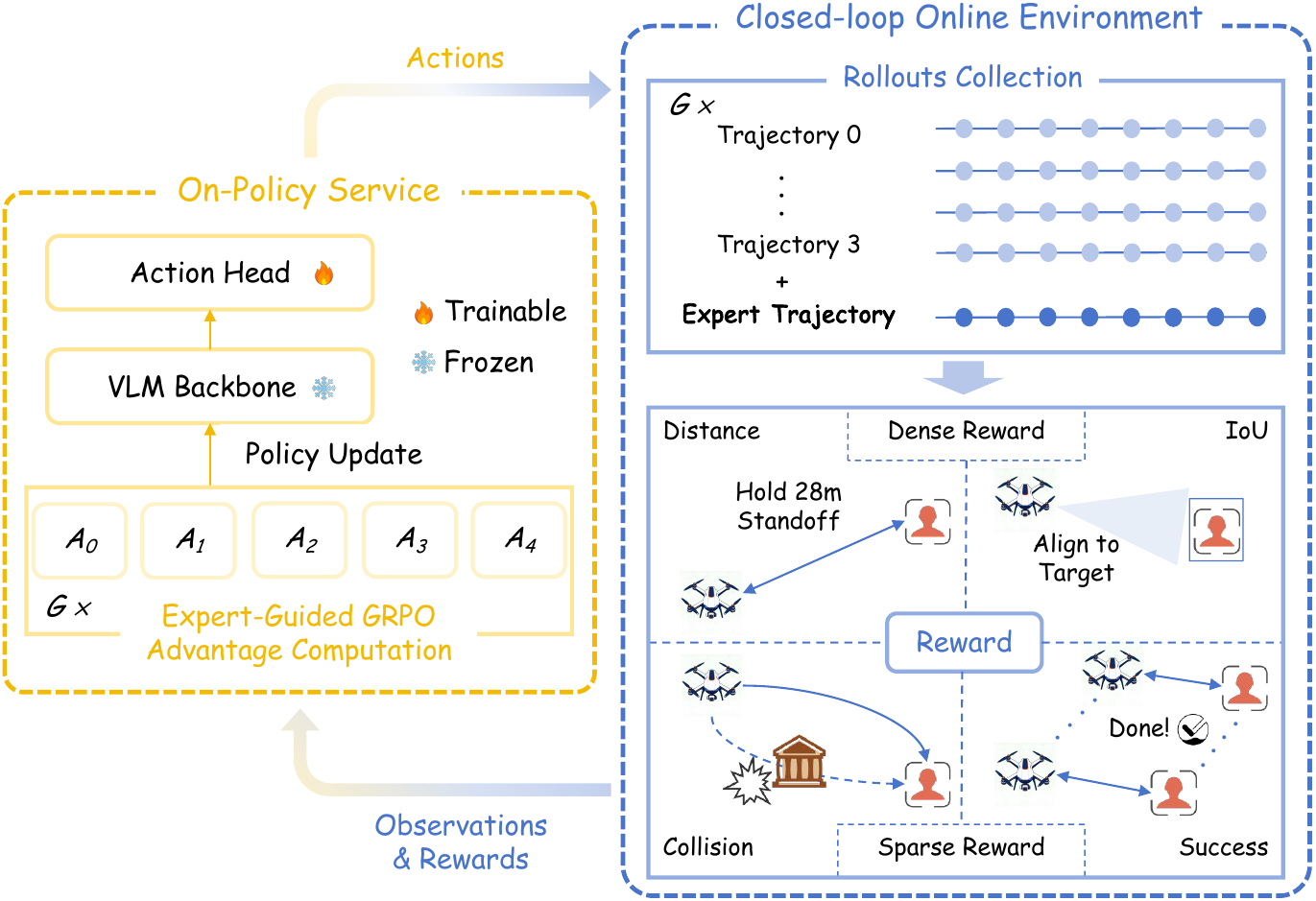}
\caption{\textbf{Closed-loop reinforcement learning pipeline.} Starting from the SFT checkpoint, CosFly-VLA is served as an action-head policy in the CARLA environment. The rollout collector gathers multiple on-policy trajectories and one expert anchor for each start condition, computes rewards from stand-off distance, target IoU, success, and collision terms, and transfers the trajectories to the trainer. The trainer computes group-relative advantages, updates the trainable action head while the backbone remains frozen, and hot-reloads the updated head for the next collection round.}
\label{fig:closed_loop_pipeline}
\end{figure}

\paragraph{Long-occlusion data mining.}
\label{sec:data_pipeline}
The SFT data pipeline starts from 6{,}785 raw Town10HD trajectories, removes 97 UAV-collision trajectories, applies target-clean filtering, and retains only trajectories with $n_{\text{bad}}<5$, where $n_{\text{bad}}$ counts frames that fail the target-validity check. This leaves 5{,}861 clean trajectories, from which JSONL generation produces 1{,}740{,}278 multimodal samples. The default window builder uses \texttt{max\_invisible=2}, which is reasonable for visible-target imitation but removes many windows needed for re-acquisition. We therefore add a relaxed miner with \texttt{max\_invisible=4}, \texttt{min\_visible\_history=0}, and \texttt{max\_alt\_jump=3.0\,m}. It surfaces 233{,}437 candidate windows, including 1{,}307 windows with 4-of-5 invisible inputs and 5{,}254 windows with 5-of-5 invisible inputs. These windows are routed into the hard and long-occlusion buckets so the model sees disappearance as a first-class training case rather than as a filtering error.
Figure~\ref{fig:closed_loop_pipeline} illustrates the rollout collection, reward computation, group-relative update, and policy hot-reload loop.

\paragraph{Closed-loop reinforcement learning.}
\label{sec:rl_stage}
The final phase targets the gap between offline supervision and closed-loop deployment, where predicted boxes and actions would be fed back into future observations and small errors can compound into drift, collision, or failure to re-acquire an occluded target. In this paper, RL optimizes the action head under action-induced observation shift, while the box/visibility history used for evaluation remains ground truth. We therefore fine-tune the SFT policy with a closed-loop reinforcement fine-tuning pipeline, denoted CosFly-RFT. The policy is initialized from the SFT checkpoint, keeps the Qwen3.5 backbone and meta-query connectors frozen, and updates the flow-matching DiT action expert through Expert-Guided Flow-Policy Optimization (EG-FPO). This design combines three well-established ideas: flow / diffusion-style continuous action generation for robot policies~\cite{lipman2023flowmatching,chi2023diffusionpolicy,black2026pi0visionlanguageactionflowmodel}, Proximal Policy Optimization (PPO)-style likelihood-ratio clipping for stable on-policy updates~\cite{schulman2017proximal}, and group-relative advantage normalization from Group Relative Policy Optimization (GRPO)~\cite{shao2024deepseekmath}.

Rollout collection is performed in a heterogeneous closed-loop system. A CARLA simulator service exposes reset and step endpoints; a rollout collector maintains the 5-frame observation window and UAV pose history; and an action-head service returns the deterministic action mean, bbox, and visibility for each state. For each path, the collector samples $K$ on-policy trajectories from a fixed-variance Gaussian proxy around the deterministic flow mean, $a\sim\mathcal{N}(\mu_\theta(s),\mathrm{diag}(\sigma^2))$. This Gaussian proxy follows the standard continuous-control policy-gradient practice used in PPO-style methods~\cite{schulman2017proximal}: the flow model supplies the mean action chunk $\mu_\theta(s)$, while the fixed diagonal variance gives a tractable log-probability for ratio-based updates without replacing the underlying flow-matching decoder. Each group is augmented with one expert-replay trajectory that acts as a high-return anchor, preventing the group advantage from collapsing when all online rollouts fail. Completed rollout JSONs are transferred to the training machine; after each update, the new action head is hot-reloaded into the serving process, and the next iteration collects rollouts with the updated policy version.

The reward is computed step by step in the simulator and combines dense tracking terms with sparse terminal events. In the main implementation, the dense terms reward maintaining the target stand-off distance and aligning the predicted bbox with the ground-truth target box, while terminal terms penalize collision and reward successful completion:
\begin{equation}
\label{eq:rl_reward}
r_t = -w_d\frac{|d_t-d^*|}{d^*} + w_{\mathrm{iou}}\mathrm{IoU}(\hat b_t,b_t^{\mathrm{GT}}) + r_t^{\mathrm{coll}} + r_t^{\mathrm{succ}},
\end{equation}
where $d^*=28$\,m, $w_d=1.0$, $w_{\mathrm{iou}}=0.5$, collision contributes a terminal penalty, and success contributes a terminal bonus. This reward is deliberately tracking-oriented: it rewards the UAV for keeping an appropriate stand-off distance, keeping the target visually grounded, avoiding obstacles, and completing the episode.

For a group $\mathcal{G}$ containing $K$ online trajectories and one expert anchor, each trajectory return is $R_i=\sum_t r_{i,t}$. Following the critic-free normalization idea in GRPO~\cite{shao2024deepseekmath}, the group-relative advantage is
\begin{equation}
\label{eq:group_advantage}
\hat A_i=\frac{R_i-\mu_{\mathcal{G}}}{\sigma_{\mathcal{G}}+\epsilon},\qquad
\mu_{\mathcal{G}}=\frac{1}{|\mathcal{G}|}\sum_{j\in\mathcal{G}}R_j,\quad
\sigma_{\mathcal{G}}=\sqrt{\frac{1}{|\mathcal{G}|}\sum_{j\in\mathcal{G}}(R_j-\mu_{\mathcal{G}})^2}.
\end{equation}
where $\epsilon$ is a small numerical stabilizer. The advantage is broadcast to every step of the corresponding trajectory. The policy ratio follows the PPO likelihood-ratio form~\cite{schulman2017proximal}, but is computed with the fixed-variance Gaussian surrogate around the flow-integrated action mean:
\begin{equation}
\label{eq:policy_ratio}
\rho_{i,t}(\theta)=\exp\!\left(\log\pi_\theta(a_{i,t}|s_{i,t})-\log\pi_{\theta_{\mathrm{old}}}(a_{i,t}|s_{i,t})\right),
\end{equation}
where $a_{i,t}$ is the sampled action chunk, $\mu_\theta(s)$ is obtained by deterministic flow integration, and $\sigma$ is recorded during rollout collection. The action expert is updated with a PPO-style clipped surrogate~\cite{schulman2017proximal},
\begin{equation}
\label{eq:eg_fpo}
\mathcal{L}_{\mathrm{EG\text{-}FPO}}(\theta)=-
\mathbb{E}_{i,t}\left[
\min\!\left(\rho_{i,t}\hat A_i,\,
\mathrm{clip}(\rho_{i,t},1-\delta,1+\delta)\hat A_i\right)
\right],
\end{equation}
with $\delta=0.2$. In practice, we also use dual clipping for negative-advantage outliers and an online-only Kullback--Leibler (KL) early stop, excluding expert steps from the KL monitor because large expert ratios are expected to pull the policy toward the anchor. A Behavior Cloning (BC) v0.1 diagnostic improved bbox IoU but caused a collision regression, so the final closed-loop RL path warm-starts directly from the SFT checkpoint. The RL split contains 550 paths across 11 maps (440 train and 110 test), plus 20 separate historical calibration paths.

%% file: chapters/05_experiments.tex
\section{Experiments}
\label{sec:exp}

\subsection{Evaluation Protocol and Metrics}
\label{sec:setup}

\begin{table}[t]
\centering
\small
\setlength{\tabcolsep}{5pt}
\caption{\textbf{Open-loop evaluation data statistics.} Held-out 5-frame windows are graded by the number of invisible frames in the four-frame history: Easy = 0, Medium = 1--2, and Hard = 3--4. Both evaluation sets are rebalanced toward occlusion relative to their natural pools, which are dominated by fully visible windows (89.7\% Easy in seen-test and 94.2\% in unseen-test); the unseen-test Medium and Hard slices retain every available window.}
\label{tab:eval_data_stats}
\begin{tabular}{l r r r r r}
\toprule
 & & & \multicolumn{3}{c}{\textbf{Occlusion slice (evaluated)}} \\
\cmidrule(lr){4-6}
\textbf{Split} & \textbf{Window pool} & \textbf{Evaluated} & \textbf{Easy} & \textbf{Medium} & \textbf{Hard} \\
\midrule
seen-test & 173{,}169 & 10{,}000 & 5{,}000 & 3{,}000 & 2{,}000 \\
unseen-test & 35{,}580 & 4{,}152 & 2{,}076 & 789 & 1{,}287 \\
\bottomrule
\end{tabular}
\end{table}

Table~\ref{tab:eval_data_stats} summarizes the open-loop evaluation pools and their occlusion-rebalanced subsets.
We evaluate CosFly-VLA in two complementary regimes that share one split convention. \emph{Open-loop} evaluation scores structured predictions on held-out 5-frame windows: the model receives identical observations and prompts, and its predicted waypoints, target box, and visibility are compared against ground truth without executing any action. \emph{Closed-loop} evaluation executes the policy in CARLA and measures end-to-end tracking outcomes; the two metric families are kept numerically separate throughout. Both regimes distinguish \textbf{seen-test}, drawn from the same map distribution as the training data, from \textbf{unseen-test}, drawn from maps disjoint from the training map distribution and therefore measuring cross-map generalization. All CosFly-VLA quantitative rows in this section use the 0.8B model; the 2B / 9B configurations are architectural scaling targets rather than evaluated rows in the present benchmark. The data composition for CPT, SFT, and RL is described in Table~\ref{tab:training_data_stats}.

\begin{table}[!t]
\centering
\scriptsize
\setlength{\tabcolsep}{2.5pt}
\renewcommand{\arraystretch}{1.2}
\setlength{\arrayrulewidth}{0.25pt}
\arrayrulecolor{black!45}
\caption{\textbf{Open-loop evaluation on seen-test.} Lower is better for ADE / DE / center distance; higher is better for IoU / Vis F1. Bold and underline mark the best and second-best values.}
\label{tab:open_loop_seen}
\resizebox{\textwidth}{!}{%
\begin{tabular}{@{}l@{\hspace{8pt}}cccc@{\hspace{10pt}}cc@{\hspace{10pt}}cc@{\hspace{10pt}}c@{}}
\toprule
\multirow{2}{*}{\textbf{Method}} & \multicolumn{4}{c}{\textbf{ADE (m)$\downarrow$}} & \multirow{2}{*}{\textbf{First DE (m)$\downarrow$}} & \multirow{2}{*}{\textbf{Final DE (m)$\downarrow$}} & \multirow{2}{*}{\textbf{IoU$\uparrow$}} & \multirow{2}{*}{\textbf{Cen. (px)$\downarrow$}} & \multirow{2}{*}{\textbf{Vis F1$\uparrow$}} \\
\cmidrule(lr){2-5}
 & \textbf{Overall} & \textbf{Easy} & \textbf{Medium} & \textbf{Hard} & & & & & \\
\midrule
\rowcolor{gray!12}\multicolumn{10}{@{}l@{}}{\emph{Closed-source VLMs}} \\
Opus-4.6~\cite{anthropic2026claudeopus46} & 1.1014 & 0.8230 & 1.3999 & 1.3499 & 0.2137 & 1.8626 & 0.033 & 81.2 & 0.939 \\
Gemini-3.1-Pro~\cite{googledeepmind2026gemini31pro} & 0.8913 & 0.7988 & 1.1616 & \textbf{0.7184} & \textbf{0.1322} & \underline{1.4106} & 0.513 & 12.3 & \textbf{0.964} \\
\rowcolor{gray!12}\multicolumn{10}{@{}l@{}}{\emph{Open-source VLMs}} \\
Qwen3-VL-235B~\cite{qwen2025qwen3vl} & 2.4861 & 2.3435 & 2.7393 & 2.4719 & 0.3955 & 4.1896 & 0.399 & 23.1 & 0.938 \\
Qwen3.5-397B~\cite{qwen2026qwen35} & 2.4508 & 2.3962 & 2.6530 & 2.2839 & 0.5054 & 4.0449 & 0.426 & 15.4 & \underline{0.952} \\
\rowcolor{gray!12}\multicolumn{10}{@{}l@{}}{\emph{General VLA models}} \\
OpenVLA~\cite{kim2024openvlaopensourcevisionlanguageactionmodel} & 1.0887 & 0.8442 & 1.3193 & 1.3542 & 0.2581 & 1.8516 & -- & -- & -- \\
$\pi_0$~\cite{black2026pi0visionlanguageactionflowmodel} & 1.2050 & 0.9066 & 1.3986 & 1.6604 & 0.3533 & 2.3197 & -- & -- & -- \\
$\pi_{0.5}$~\cite{intelligence2025pi05visionlanguageactionmodelopenworld} & 1.1407 & 0.8690 & 1.3301 & 1.5357 & 0.3318 & 2.1766 & -- & -- & -- \\
\midrule
\textbf{CosFly-VLA-0.8B (SFT)} & \underline{0.8247} & \underline{0.6719} & \underline{0.9893} & 0.9599 & 0.2150 & 1.4750 & \underline{0.649} & \underline{9.7} & 0.938 \\
\textbf{CosFly-VLA-0.8B (SFT+CoT)} & \textbf{0.7175} & \textbf{0.6351} & \textbf{0.8522} & \underline{0.7218} & \underline{0.1884} & \textbf{1.2929} & \textbf{0.659} & \textbf{9.4} & 0.943 \\
\bottomrule
\end{tabular}%
}
\vspace{0.55em}
\caption{\textbf{Open-loop evaluation on unseen-test.} This split evaluates cross-map generalization. Bold and underline mark the best and second-best values.}
\label{tab:open_loop_unseen}
\resizebox{\textwidth}{!}{%
\begin{tabular}{@{}l@{\hspace{8pt}}cccc@{\hspace{10pt}}cc@{\hspace{10pt}}cc@{\hspace{10pt}}c@{}}
\toprule
\multirow{2}{*}{\textbf{Method}} & \multicolumn{4}{c}{\textbf{ADE (m)$\downarrow$}} & \multirow{2}{*}{\textbf{First DE (m)$\downarrow$}} & \multirow{2}{*}{\textbf{Final DE (m)$\downarrow$}} & \multirow{2}{*}{\textbf{IoU$\uparrow$}} & \multirow{2}{*}{\textbf{Cen. (px)$\downarrow$}} & \multirow{2}{*}{\textbf{Vis F1$\uparrow$}} \\
\cmidrule(lr){2-5}
 & \textbf{Overall} & \textbf{Easy} & \textbf{Medium} & \textbf{Hard} & & & & & \\
\midrule
\rowcolor{gray!12}\multicolumn{10}{@{}l@{}}{\emph{Closed-source VLMs}} \\
Opus-4.6~\cite{anthropic2026claudeopus46} & 1.0232 & 0.5015 & 1.1021 & 1.8162 & 0.2278 & 1.8725 & 0.023 & 97.3 & 0.806 \\
Gemini-3.1-Pro~\cite{googledeepmind2026gemini31pro} & 0.6442 & \textbf{0.4560} & 1.0690 & \textbf{0.6823} & \textbf{0.1300} & 1.2587 & 0.516 & 14.2 & \underline{0.929} \\
\rowcolor{gray!12}\multicolumn{10}{@{}l@{}}{\emph{Open-source VLMs}} \\
Qwen3-VL-235B~\cite{qwen2025qwen3vl} & 2.2616 & 2.1803 & 2.3913 & 2.2992 & 0.3913 & 4.1532 & 0.361 & 26.5 & 0.852 \\
Qwen3.5-397B~\cite{qwen2026qwen35} & 2.1319 & 2.1036 & 2.1891 & 2.1425 & 0.4325 & 3.9331 & 0.428 & 19.5 & 0.874 \\
\rowcolor{gray!12}\multicolumn{10}{@{}l@{}}{\emph{General VLA models}} \\
OpenVLA~\cite{kim2024openvlaopensourcevisionlanguageactionmodel} & 0.9160 & 0.6715 & 1.1500 & 1.1671 & 0.2253 & 1.6128 & -- & -- & -- \\
$\pi_0$~\cite{black2026pi0visionlanguageactionflowmodel} & 1.5611 & 1.4001 & 1.6036 & 1.7946 & 0.3825 & 2.7218 & -- & -- & -- \\
$\pi_{0.5}$~\cite{intelligence2025pi05visionlanguageactionmodelopenworld} & 1.4279 & 1.1733 & 1.5364 & 1.7723 & 0.3584 & 2.4295 & -- & -- & -- \\
\midrule
\textbf{CosFly-VLA-0.8B (SFT)} & \underline{0.6364} & 0.4874 & \underline{0.7550} & 0.8039 & 0.1610 & \underline{1.1550} & \underline{0.616} & \underline{12.5} & 0.878 \\
\textbf{CosFly-VLA-0.8B (SFT+CoT)} & \textbf{0.5931} & \underline{0.4782} & \textbf{0.7054} & \underline{0.7096} & \underline{0.1413} & \textbf{1.1257} & \textbf{0.635} & \textbf{11.4} & \textbf{0.932} \\
\bottomrule
\end{tabular}%
}
\arrayrulecolor{black}
\end{table}

\paragraph{Open-loop evaluation data.} Open-loop difficulty is defined at the 5-frame window level, not at the episode level used for closed-loop evaluation in Section~\ref{sec:task}. Let $v_j\in\{0,1\}$ denote the simulator visibility flag at frame $j$, where one indicates visible. For a sample whose prediction is made at time $t$, let
$n_{\mathrm{inv}}^{\mathrm{hist}}=\sum_{j=1}^{4}\mathbf{1}[v_{t-j}=0]$
be the number of invisible frames in the four historical input frames. We define Easy as $n_{\mathrm{inv}}^{\mathrm{hist}}=0$, Medium as $n_{\mathrm{inv}}^{\mathrm{hist}}\in\{1,2\}$, and Hard as $n_{\mathrm{inv}}^{\mathrm{hist}}\in\{3,4\}$. The natural window distribution is dominated by fully visible targets (89.7\% of the seen-test pool and 94.2\% of the unseen-test pool are Easy), so uniform sampling would mostly measure visible-frame tracking. We therefore evaluate on occlusion-rebalanced sets: 10{,}000 seen-test windows (5{,}000 Easy / 3{,}000 Medium / 2{,}000 Hard) drawn from a 173{,}169-window pool, and 4{,}152 unseen-test windows (2{,}076 Easy / 789 Medium / 1{,}287 Hard) drawn from a 35{,}580-window pool, where the unseen-test Medium and Hard slices retain every available window.

\paragraph{Metrics.} \emph{Open-loop:} let $\hat p_{t+k}$ and $p_{t+k}$ denote the predicted and ground-truth 3-D waypoint positions implied by the 8-step action chunk. We report Average Displacement Error (ADE) as $\frac{1}{H_a}\sum_{k=1}^{H_a}\|\hat p_{t+k}-p_{t+k}\|_2$, First Displacement Error (First DE) as $\|\hat p_{t+1}-p_{t+1}\|_2$, and Final Displacement Error (Final DE, equivalently FDE) as $\|\hat p_{t+H_a}-p_{t+H_a}\|_2$, all in meters. For grounding, bbox Intersection-over-Union (IoU) is computed between the predicted and ground-truth current-frame boxes, and center error (\emph{Cen.}) is the Euclidean pixel distance between their centers after mapping Qwen-style coordinates back to the $640{\times}360$ image. Visibility F1 (Vis F1) treats visible as the positive class and thresholds the predicted visibility probability at $0.5$. \emph{Closed-loop:} Success Rate (SR) counts an episode as successful when its visible-frame rate is at least $0.8$, it has no collision, and the final five rollout steps are visible. Track Continuity Ratio (TCR) is the longest contiguous visible segment divided by the episode length. Following the benchmark convention, Average Tracking Frames (ATF) is the mean number of valid tracking decision steps before the first fatal failure, defined as either a collision or two consecutive invisible steps; one decision step spans five simulator frames ($2.5$\,s). Closed-loop ADE is the mean Euclidean error between the UAV rollout trajectory and the expert / ground-truth tracking trajectory, and stand-off distance error (dErr) is the mean $|d_t-d^*|$ with $d^*=28$\,m.

\paragraph{Implementation and training configuration.}
All training stages reported in this paper---CPT, multi-stage SFT, CoT training, and closed-loop RL---were run on a single node with eight NVIDIA A800 GPUs. The SFT phase of CosFly-VLA-0.8B starts from the spatially grounded CPT backbone (full-parameter CPT including the ViT) and follows the three-stage curriculum of Section~\ref{sec:curriculum}; All stages share the same architecture configuration: a 12-layer, 8-head cross-attention DiT action expert over an $8{\times}4$ action chunk $(\Delta x,\Delta y,\Delta z,\Delta\psi)$ with per-dimension normalization scales $(5,5,2,30)$, meta-query pooling with 16 action, 16 bbox, and 1 visibility queries, a 2-layer meta-query connector, and 8 flow-integration steps at inference. Optimization uses bf16 with DeepSpeed ZeRO-2, AdamW without weight decay, a cosine learning-rate schedule. The effective batch size is fixed at 32 throughout (8 GPUs $\times$ 4 per-GPU batch $\times$ 1 gradient accumulation).

\subsection{Open-loop Evaluation}
\label{sec:internal_eval}

Open-loop evaluation measures, on held-out 5-frame windows, whether a model can parse the task prompt and predict plausible waypoints, boxes, and visibility before closed-loop feedback compounds errors. We use it to position CosFly-VLA against two reference groups under the \emph{same} prompt and I/O contract: (i) general-purpose VLMs evaluated \emph{zero-shot}, split into closed-source and open-source models; and (ii) general VLA baselines, including OpenVLA, $\pi_0$, and $\pi_{0.5}$. Because the general VLMs have no calibrated action head, their displacement errors largely reflect prompt-only spatial extrapolation. The VLA baselines are therefore the primary reference group for action prediction, while the zero-shot VLMs contextualize the difficulty of structured waypoint regression. Table~\ref{tab:open_loop_seen} reports the same-map seen-test comparison, while Table~\ref{tab:open_loop_unseen} focuses on cross-map generalization under unseen-test. Both tables are structured-output comparisons only and are never compared numerically with closed-loop rollout metrics.

Among the VLA baselines, CosFly-VLA-0.8B (SFT+CoT) achieves the lowest waypoint error on both splits and all occlusion slices. Relative to OpenVLA, it reduces overall ADE by 34.1\% on seen-test (1.0887 $\rightarrow$ 0.7175) and by 35.3\% on unseen-test (0.9160 $\rightarrow$ 0.5931), while reducing Final DE by about 30\% on both splits. The improvement is most visible in hard cases: Hard ADE drops by 46.7\% on seen-test and 39.2\% on unseen-test relative to OpenVLA. The zero-shot VLMs remain competitive in several short-horizon columns, especially First DE, but CosFly-VLA provides the best matched bbox grounding among evaluated models with bbox outputs, reaching 0.659 / 0.635 IoU and 9.4 / 11.4 px center error on seen-test / unseen-test. This stronger grounding may help preserve target identity under occlusion, although the open-loop metrics alone do not establish a causal reduction in identity switches.

\begin{table}[t]
\centering
\footnotesize
\setlength{\tabcolsep}{2.6pt}
\renewcommand{\arraystretch}{1.08}
\caption{\textbf{Closed-loop comparison on seen-test and unseen-test.} SR, TCR, and ATF are higher-is-better; ADE and dErr are lower-is-better. ATF is reported in decision steps, while ADE and dErr are in meters. Bold and underline mark the best and second-best values.}
\label{tab:closed_loop_sota}
\begin{tabular}{@{}l@{\hspace{5pt}}cc@{\hspace{7pt}}cc@{\hspace{7pt}}cc@{\hspace{7pt}}cc@{\hspace{7pt}}cc@{}}
\toprule
\multirow{2}{*}{\textbf{Method}} & \multicolumn{2}{c}{\textbf{SR$\uparrow$}} & \multicolumn{2}{c}{\textbf{TCR$\uparrow$}} & \multicolumn{2}{c}{\textbf{ATF$\uparrow$}} & \multicolumn{2}{c}{\textbf{ADE$\downarrow$}} & \multicolumn{2}{c}{\textbf{dErr$\downarrow$}} \\
\cmidrule(lr){2-3}\cmidrule(lr){4-5}\cmidrule(lr){6-7}\cmidrule(lr){8-9}\cmidrule(lr){10-11}
 & \textbf{Seen} & \textbf{Unseen} & \textbf{Seen} & \textbf{Unseen} & \textbf{Seen} & \textbf{Unseen} & \textbf{Seen} & \textbf{Unseen} & \textbf{Seen} & \textbf{Unseen} \\
\midrule
\rowcolor{gray!12}\multicolumn{11}{@{}l@{}}{\emph{General VLA models}} \\
$\pi_0$~\cite{black2026pi0visionlanguageactionflowmodel} & 41\% & 47\% & 68\% & 59\% & 25.4 & 19.4 & 14.4 & 17.6 & 8.7 & 10.4 \\
$\pi_{0.5}$~\cite{intelligence2025pi05visionlanguageactionmodelopenworld} & 47\% & 60\% & 68\% & 64\% & 27.3 & 21.5 & 12.9 & 14.0 & 6.3 & 7.3 \\
OpenVLA~\cite{kim2024openvlaopensourcevisionlanguageactionmodel} & 57\% & \underline{80\%} & 73\% & 70\% & 28.4 & \underline{22.8} & 9.0 & \underline{6.9} & 3.4 & \underline{3.3} \\
\midrule
\textbf{CosFly-VLA-0.8B (SFT)} & \underline{70\%} & 79\% & \textbf{79\%} & \textbf{72\%} & \textbf{30.5} & \underline{22.8} & \underline{8.9} & 8.8 & \underline{3.3} & 3.9 \\
\textbf{CosFly-VLA-0.8B (RL)} & \textbf{74\%} & \textbf{82\%} & \underline{74\%} & \underline{71\%} & \underline{29.6} & \textbf{23.5} & \textbf{8.1} & \textbf{6.3} & \textbf{2.7} & \textbf{2.7} \\
\bottomrule
\end{tabular}
\end{table}

\begin{table}[t]
\centering
\scriptsize
\setlength{\tabcolsep}{3pt}
\caption{\textbf{Training-recipe ablation on open-loop ADE.} All rows use the same 0.8B model scale and are evaluated on the seen-test / unseen-test splits in Table~\ref{tab:eval_data_stats}. Lower is better; bold marks the best value and underline marks the second-best value within each split and occlusion slice.}
\label{tab:open_loop_ablation}
\resizebox{\textwidth}{!}{%
\begin{tabular}{@{}l c c c r r r r r r r r@{}}
\toprule
\multirow{2}{*}{\textbf{Variant}} & \multirow{2}{*}{\textbf{Curr.}} & \multirow{2}{*}{\textbf{CPT}} & \multirow{2}{*}{\textbf{CoT}} & \multicolumn{4}{c}{\textbf{seen-test ADE (m)$\downarrow$}} & \multicolumn{4}{c}{\textbf{unseen-test ADE (m)$\downarrow$}} \\
\cmidrule(lr){5-8}\cmidrule(lr){9-12}
 & & & & \textbf{Overall} & \textbf{Easy} & \textbf{Medium} & \textbf{Hard} & \textbf{Overall} & \textbf{Easy} & \textbf{Medium} & \textbf{Hard} \\
\midrule
SFT only & \xmark & \xmark & \xmark & 0.8924 & 0.7265 & 1.0671 & 1.0451 & 0.7087 & 0.5490 & 0.7815 & 0.9217 \\
\quad + curriculum learning & \cmark & \xmark & \xmark & 0.8295 & 0.6895 & \underline{0.9711} & 0.9670 & 0.7007 & 0.5301 & 0.7782 & 0.9284 \\
\quad + spatially grounded CPT & \cmark & \cmark & \xmark & \underline{0.8247} & \underline{0.6719} & 0.9893 & \underline{0.9599} & \underline{0.6364} & \underline{0.4874} & \underline{0.7550} & \underline{0.8039} \\
\quad + CoT supervision & \cmark & \cmark & \cmark & \textbf{0.7175} & \textbf{0.6351} & \textbf{0.8522} & \textbf{0.7218} & \textbf{0.5931} & \textbf{0.4782} & \textbf{0.7054} & \textbf{0.7096} \\
\bottomrule
\end{tabular}%
}
\end{table}

\subsection{Closed-loop Evaluation}
\label{sec:main_results}

Closed-loop evaluation executes the predicted action chunks and feeds the action-induced observations back into the policy. The seen-test set contains 100 new Town10HD\_Opt episodes (34 Easy / 33 Medium / 33 Hard), while unseen-test contains 90 episodes from Town01\_Opt, Town03\_Opt, and Town05\_Opt (30 per town, with 10 episodes per difficulty grade). For a controlled comparison, all policies use deterministic mean actions, simulator collision checking, and the same five-frame ground-truth history buffer. Table~\ref{tab:closed_loop_sota} reports SR, TCR, ATF, rollout ADE, and stand-off distance error. These numbers measure behavior after action and trajectory errors have had a chance to compound.

The RL-tuned CosFly-VLA-0.8B achieves the best SR, rollout ADE, and dErr on both splits, but it is not uniformly best on every closed-loop continuity metric. Compared with the strongest general VLA baseline, OpenVLA, it improves SR by 17 percentage points on seen-test (57\% $\rightarrow$ 74\%, a 29.8\% relative increase) and by 2 points on unseen-test (80\% $\rightarrow$ 82\%, a 2.5\% relative increase). Rollout ADE drops by 10.0\% on seen-test (9.0 $\rightarrow$ 8.1) and 8.7\% on unseen-test (6.9 $\rightarrow$ 6.3), while stand-off distance error decreases by 20.6\% and 18.2\%, respectively. TCR improves modestly over OpenVLA (73\% $\rightarrow$ 74\% on seen-test and 70\% $\rightarrow$ 71\% on unseen-test), and ATF improves by 4.2\% / 3.1\% on seen-test / unseen-test. Relative to the SFT checkpoint, RL improves SR by 5.7\% / 3.8\% and reduces ADE by 9.0\% / 28.4\% on seen-test / unseen-test, but the SFT model retains slightly higher TCR and seen-test ATF. This suggests that the RL update primarily improves success, trajectory accuracy, and stand-off control.

\begin{figure}[t]
\centering
\includegraphics[width=0.8\linewidth]{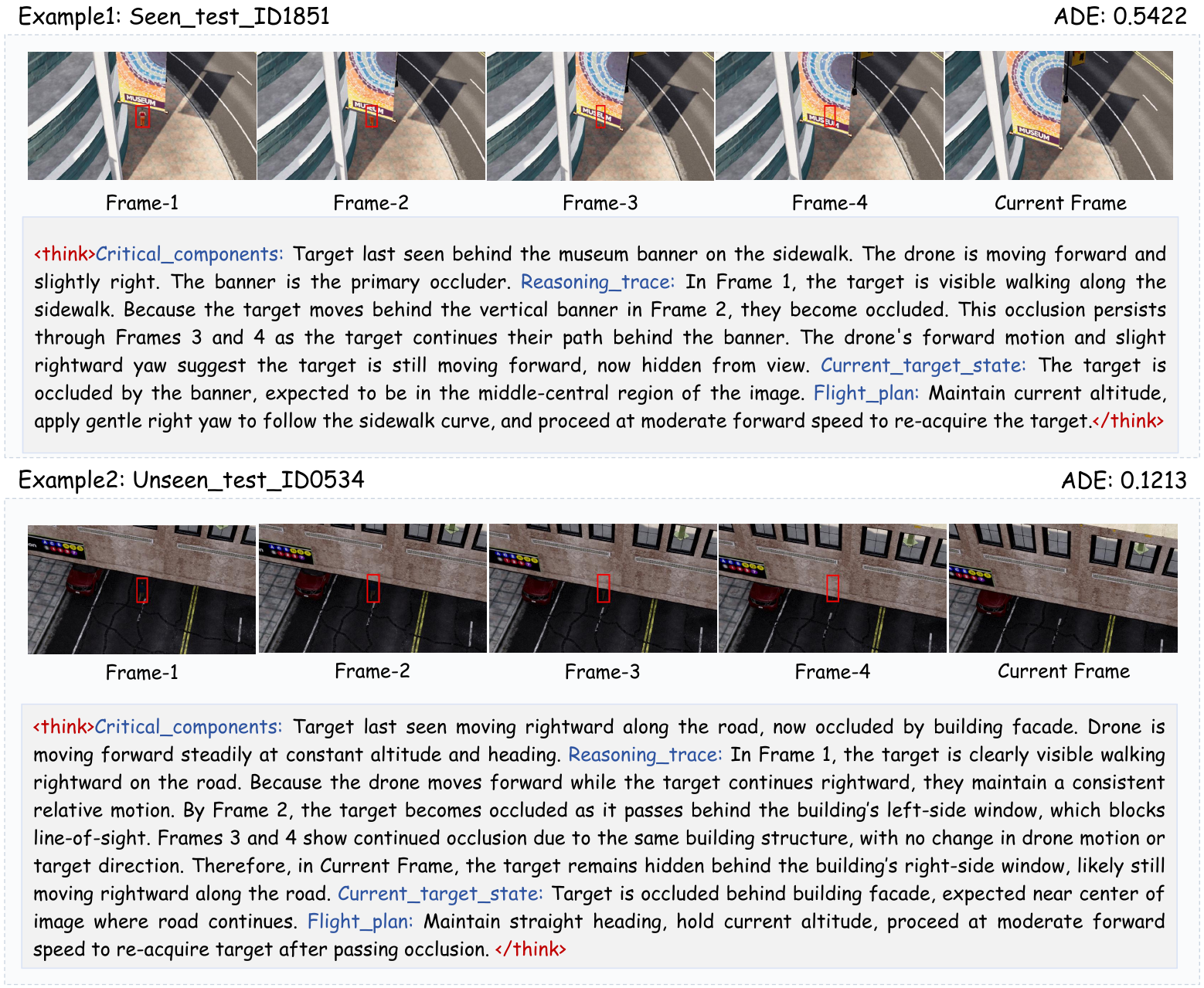}
\caption{\textbf{Qualitative CoT reasoning under long occlusion.} The seen-test and unseen-test examples each show four historical frames and the current frame, followed by a structured reasoning trace that identifies the last visible target state, the primary occluder, relative motion, the estimated current target region, and a recovery-oriented flight plan.}
\label{fig:cot_example}
\end{figure}

\subsection{Ablation Studies}
\label{sec:ablation}

Table~\ref{tab:open_loop_ablation} reports a cumulative ablation of the open-loop training recipe. Starting from an SFT-only baseline, we progressively add curriculum learning, spatially grounded CPT, and CoT supervision while keeping the model scale and evaluation protocol fixed. This design isolates how each component changes structured waypoint prediction.

The cumulative ablation shows that the gains are not confined to easy visible-frame tracking. Adding curriculum learning reduces seen-test overall ADE by 7.0\% relative to SFT only (0.8924 $\rightarrow$ 0.8295). The step that adds spatially grounded CPT coincides with a 9.2\% reduction in unseen-test overall ADE (0.7007 $\rightarrow$ 0.6364) and a 13.4\% reduction in unseen-test Hard ADE (0.9284 $\rightarrow$ 0.8039). The subsequent CoT stage yields the largest marginal gain on seen-test Hard, reducing ADE by 24.8\% (0.9599 $\rightarrow$ 0.7218), and further reduces unseen-test Hard ADE by 11.7\% (0.8039 $\rightarrow$ 0.7096). From the SFT-only baseline to the final $+$CoT variant, overall ADE decreases by 19.6\% on seen-test and 16.3\% on unseen-test, with Hard ADE decreasing by 30.9\% and 23.0\%, respectively.

\begin{figure}[H]
\centering
\includegraphics[width=1.0\linewidth]{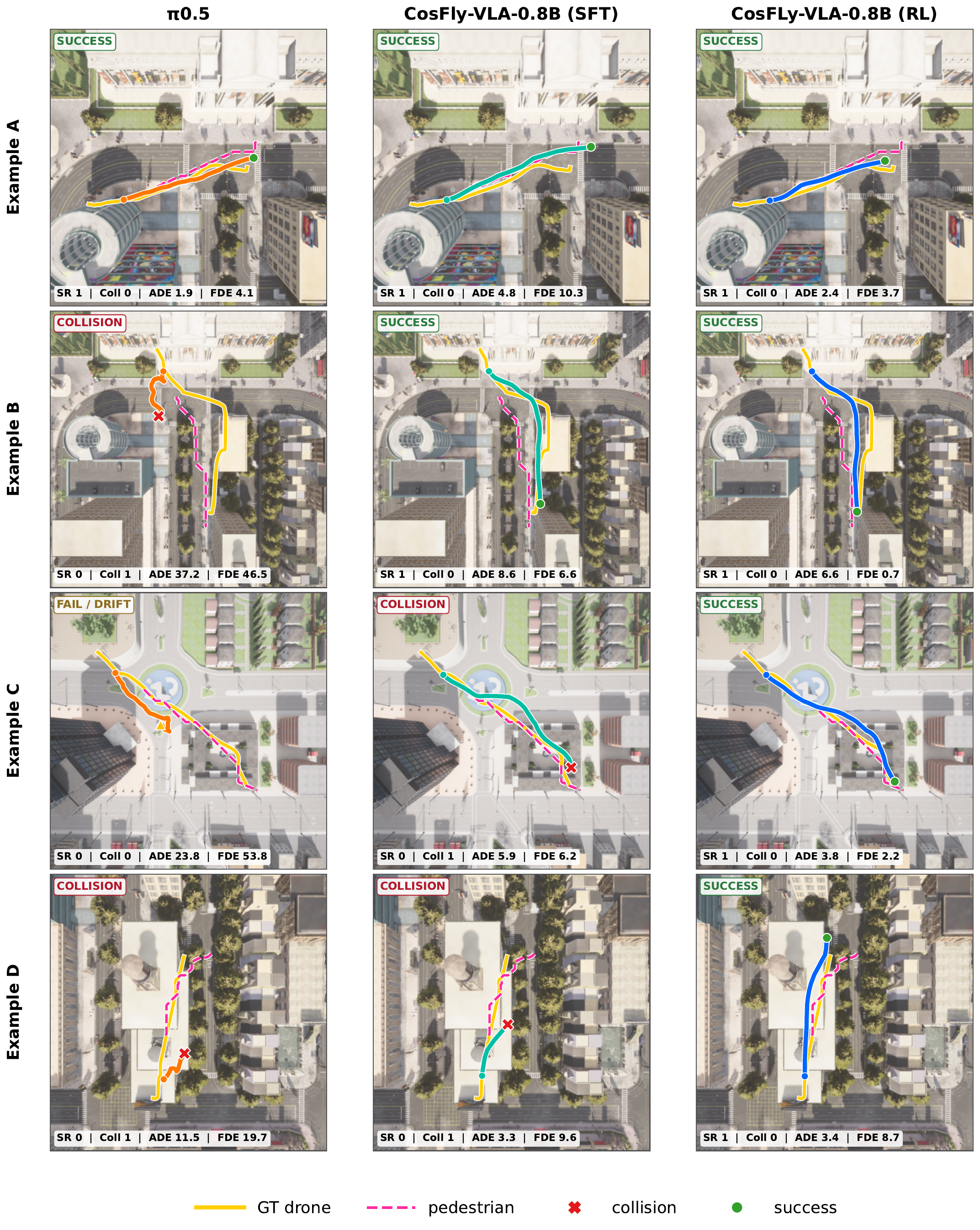}
\caption{\textbf{Qualitative closed-loop comparison on representative CARLA episodes.} Rows show Examples~A--D; columns compare $\pi_{0.5}$, CosFly-VLA-0.8B (SFT), and CosFly-VLA-0.8B (RL). The thick bright-yellow curve denotes the ground-truth UAV path, and the thick bright-magenta dashed curve denotes pedestrian motion. Executed UAV trajectories are shown as thick solid curves, using vivid orange for $\pi_{0.5}$, bright cyan-green for CosFly-VLA-0.8B (SFT), and bright blue for CosFly-VLA-0.8B (RL). Markers indicate success, collision, or drift, and each panel reports success, collision count, ADE, and FDE.}
\label{fig:qualitative_analysis}
\end{figure}

\subsection{Chain-of-Thought Reasoning for Hard Scenes}
\label{sec:cot}

Hard open-loop windows contain three or four invisible history frames, so the model must reason about where the target went and where it may re-emerge rather than track appearance alone. The quantitative change associated with the CoT step is reported in Table~\ref{tab:open_loop_ablation}: it produces its largest marginal improvement on seen-test Hard (0.9599 $\rightarrow$ 0.7218) and also improves unseen-test Hard (0.8039 $\rightarrow$ 0.7096). Figure~\ref{fig:cot_example} complements these aggregate results with one seen-test and one unseen-test example.

In the seen-test example, the trace identifies a banner as the primary occluder and extrapolates the pedestrian's motion along the sidewalk; in the unseen-test example, it attributes the disappearance to a building facade and preserves the target's rightward motion hypothesis. In both cases, the model explicitly marks the target as occluded rather than treating missing appearance evidence as a new target, then converts the inferred state into a conservative forward / yaw command intended to restore line of sight. These examples provide qualitative evidence of the reasoning pathway; the corresponding ADE changes are reported by the cumulative ablation in Table~\ref{tab:open_loop_ablation}.

\subsection{Qualitative Visualization Analysis}
\label{sec:qualitative_analysis}

The aggregate metrics demonstrate that closed-loop reinforcement learning improves both tracking success and flight safety, but they do not explain how these improvements emerge in individual environments. We therefore visualize four representative closed-loop episodes in Figure~\ref{fig:qualitative_analysis}. Each row presents the same CARLA scenario for three methods: the general-purpose VLA baseline $\pi_{0.5}$, CosFly-VLA-0.8B (SFT), and CosFly-VLA-0.8B (RL). The trajectories are overlaid on the actual CARLA map so that prediction drift and collision locations can be interpreted relative to buildings, vegetation, intersections, and narrow streets.

Example~A provides a sanity case in which all three methods complete the episode successfully. This case shows that CosFly-VLA-0.8B (RL) preserves stable tracking behavior in ordinary scenarios rather than improving safety by adopting an excessively conservative policy. In Example~B, $\pi_{0.5}$ accumulates substantial closed-loop error and collides, whereas both CosFly variants complete the route; the RL policy additionally reduces the terminal trajectory error.

The remaining examples highlight more challenging failure modes. In Example~C, $\pi_{0.5}$ drifts away from the target trajectory, while CosFly-VLA-0.8B (SFT) follows the reference more closely but eventually collides near the surrounding structures. CosFly-VLA-0.8B (RL) instead maintains a collision-free trajectory and successfully completes the episode. Examples~D is stronger safety-critical cases: both $\pi_{0.5}$ and CosFly-VLA-0.8B (SFT) collide in dense urban corridors containing buildings and vegetation, whereas CosFly-VLA-0.8B (RL) reaches the end safely. Notably, the RL trajectory may locally deviate from the oracle path to preserve clearance. This illustrates why trajectory error alone is insufficient for closed-loop evaluation: a path with slightly lower instantaneous ADE may still be unusable if it terminates in collision.
These qualitative patterns agree with the quantitative results in Table~\ref{tab:closed_loop_sota}. On seen-test, CosFly-VLA-0.8B (RL) improves SR from $47\%$ for $\pi_{0.5}$ and $70\%$ for CosFly-VLA-0.8B (SFT) to $74\%$. On unseen-test, it reaches an SR of $82\%$ and yields the lowest rollout ADE and stand-off distance error among the compared methods. Together, the visualizations indicate that the benefit of closed-loop RL is not limited to more accurate waypoint regression; it produces trajectories that remain executable, maintain target tracking, and avoid safety-critical obstacles under compounding closed-loop errors.

%% file: chapters/06_discussion.tex
\section{Discussion}
\label{sec:discussion}

Taken together, the open-loop seen-test / unseen-test comparisons, training-recipe ablations, closed-loop rollouts, and diagnostic cases support two conclusions. First, the 0.8B SFT+CoT variant improves structured waypoint prediction over VLA baselines and over its SFT-only baseline and intermediate variants. Second, the RL-tuned 0.8B policy improves SR, rollout ADE, and stand-off distance error over the strongest general VLA baseline with matched logs. These conclusions apply to the aggregate split-level metrics defined by the stated evaluation protocol.

The most important lesson is that UAV tracking should be treated as recovery under partial observability, not only as visible-frame localization. The open-loop Hard slices and closed-loop rollout examples both stress cases where target evidence becomes unreliable. Figure~\ref{fig:cot_example} shows how the policy maintains an explicit hypothesis about the unseen target, while Figure~\ref{fig:qualitative_analysis} shows how closed-loop errors can develop into drift or collision. Together, these results suggest that robust tracking depends on maintaining a useful belief about an unseen target and preventing the agent's own outputs from becoming a self-reinforcing failure mode.

The second lesson is that data construction and training schedule are part of the method. A default window builder that filters out 4-of-5 and 5-of-5 invisible-input windows makes the dataset look cleaner, but it also removes the episodes most relevant to re-acquisition. The relaxed miner and Stage-3 long-occlusion quota are therefore not merely implementation choices; they encode the failure distribution that the policy must learn. The same logic applies to spatially grounded CPT: the point is not just to add more pretraining data, but to expose the frozen Qwen3.5 backbone to the aerial geometry cues that remain informative when the target is absent.

The final lesson is systems-level. Closed-loop tracking quality is shaped by serving latency and feedback design as much as by offline prediction quality; the current simulator-side evaluation does not establish high-frequency onboard readiness. Likewise, the difference between the SFT and RL rows in Table~\ref{tab:closed_loop_sota} confirms that improvements in static structured outputs do not automatically imply the best on-policy behavior, because executed actions change the subsequent observations and state distribution. Multi-seed evaluation, predicted-box history feedback, per-occlusion closed-loop analysis, and eventual sim-to-real validation are natural extensions beyond the present benchmark.

%% file: chapters/07_limitations_and_futurework.tex
\section{Limitations and Future Work}
\label{sec:limitations}

CosFly-VLA still has several limitations. First, the current training data is mainly collected from a limited set of simulated environments, maps, and target categories. This setting is useful for controlled method development, but it cannot cover the long-tail diversity of real UAV deployment, where road layouts, building structures, camera viewpoints, illumination, weather, and target appearances vary substantially. Second, target behavior in the present data remains relatively simple: most targets move with limited state variation, often at near-constant velocity, and the scenes are primarily outdoor. More challenging cases, such as outdoor-to-indoor transitions, dense crowds, abrupt intent changes, and adversarial motion, are not yet well represented. Third, the reported quantitative comparisons use the 0.8B model and aggregate split-level metrics; they do not establish one-to-one scaling to the 2B / 9B configurations or quantify multi-seed variance. The closed-loop benchmark also uses a shared ground-truth history buffer to isolate action-policy behavior, so robustness to predicted-box history feedback remains outside the present evaluation. Finally, all evaluations are conducted in simulation. The results therefore do not establish robustness under physical UAV dynamics, onboard sensing noise, communication latency, real occluders, or safety constraints in deployment.

These limitations point to three directions for future work. The first is to scale model size, environmental diversity, and target-behavior diversity, including more CARLA towns, richer object categories, variable target speeds, stop-and-go behavior, abrupt turns, indoor--outdoor transitions, and more realistic long-occlusion events. The second is to extend closed-loop evaluation with multiple seeds, per-occlusion reporting, and predicted-box history feedback, while developing reinforcement-learning methods that learn from harder on-policy failures and explicitly optimize re-acquisition. The third direction is sim-to-real transfer and physical deployment. Evaluating CosFly-VLA in AirSim or other simulators, adapting the policy to real sensor noise and control latency, and eventually testing on physical UAV platforms will be necessary to determine whether spatially grounded VLA tracking can support reliable target recovery beyond simulation.

%% file: chapters/08_conclusion.tex
\section{Conclusion}
\label{sec:conclusion}
In this work, we presented \textbf{CosFly-VLA}, a spatially aware vision-language-action framework for occlusion-robust UAV target tracking. Rather than treating aerial tracking as visible-frame localization followed by low-level control, CosFly-VLA formulates the task as closed-loop target recovery: the policy must ground the target when it is visible, reason about its likely re-emergence when it is occluded, and correct action-level errors induced by its own previous motions. Establishing the same recovery behavior under predicted-state feedback, rather than the ground-truth state history used here, remains future work. To bridge language-conditioned perception and continuous UAV control, we introduced a meta-query interface with decoupled heads for waypoint prediction, target-box grounding, and visibility estimation. We further developed a training recipe that combines Spatially Grounded Continued Pretraining, three-stage Curriculum-based Supervised Fine-Tuning, Chain-of-Thought training, and closed-loop Reinforcement Learning guided by tracking-oriented rewards. Relative to OpenVLA, CosFly-VLA-0.8B (SFT+CoT) reduces open-loop overall ADE by 34.1\% / 35.3\% on seen-test / unseen-test; the cumulative ablation associates the strongest cross-map gain with adding spatially grounded CPT and the largest seen-test Hard gain with adding CoT supervision. Also relative to OpenVLA, the RL-tuned variant improves closed-loop SR by 29.8\% (+17 percentage points) on seen-test and 2.5\% (+2 points) on unseen-test, while reducing rollout ADE by 10.0\% / 8.7\% and stand-off distance error by 20.6\% / 18.2\%. These simulation results provide evidence that UAV VLA systems can move from visible-frame imitation toward spatially grounded action-closed-loop control under a shared ground-truth state history. Beyond the present benchmark, future work may study multi-seed confidence intervals, per-occlusion closed-loop behavior, and physical UAV deployment.

%% file: references.bib
@misc{zhao2023learningfinegrainedbimanualmanipulation,
      title={Learning Fine-Grained Bimanual Manipulation with Low-Cost Hardware},
      author={Tony Z. Zhao and Vikash Kumar and Sergey Levine and Chelsea Finn},
      year={2023},
      eprint={2304.13705},
      archivePrefix={arXiv},
      primaryClass={cs.RO},
      url={https://arxiv.org/abs/2304.13705},
}

@inproceedings{jiang2025survey,
  title={A Survey on Vision-Language-Action Models for Autonomous Driving},
  author={Jiang, Sicong and Huang, Zilin and Qian, Kangan and Luo, Ziang and Zhu, Tianze and Zhong, Yang and Tang, Yihong and Kong, Menglin and Wang, Yunlong and Jiao, Siwen and others},
  booktitle={Proceedings of the IEEE/CVF International Conference on Computer Vision (ICCV) Workshops},
  pages={4583--4595},
  year={2025},
  eprint={2506.24044},
  archivePrefix={arXiv},
  primaryClass={cs.RO},
  url={https://arxiv.org/abs/2506.24044}
}

@misc{zhang2025uninavidvideobasedvisionlanguageactionmodel,
      title={Uni-NaVid: A Video-based Vision-Language-Action Model for Unifying Embodied Navigation Tasks},
      author={Jiazhao Zhang and Kunyu Wang and Shaoan Wang and Minghan Li and Haoran Liu and Songlin Wei and Zhongyuan Wang and Zhizheng Zhang and He Wang},
      year={2025},
      eprint={2412.06224},
      archivePrefix={arXiv},
      primaryClass={cs.RO},
      url={https://arxiv.org/abs/2412.06224},
}

@misc{wu2025hierarchicalinstructionawareembodiedvisual,
      title={Hierarchical Instruction-aware Embodied Visual Tracking},
      author={Kui Wu and Hao Chen and Churan Wang and Fakhri Karray and Zhoujun Li and Yizhou Wang and Fangwei Zhong},
      year={2025},
      eprint={2505.20710},
      archivePrefix={arXiv},
      primaryClass={cs.CV},
      url={https://arxiv.org/abs/2505.20710},
}

@inproceedings{11246600,
  author={Wu, Kui and Xu, Shuhang and Chen, Hao and Wang, Churan and Li, Zhoujun and Wang, Yizhou and Zhong, Fangwei},
  booktitle={2025 IEEE/RSJ International Conference on Intelligent Robots and Systems (IROS)},
  title={{VLM} Can Be a Good Assistant: Enhancing Embodied Visual Tracking with Self-Improving Vision-Language Models},
  year={2025},
  pages={13154--13161},
  doi={10.1109/IROS60139.2025.11246600},
  url={https://doi.org/10.1109/IROS60139.2025.11246600}
}

@misc{liu2025trackvlaunleashingreasoningmemory,
      title={TrackVLA++: Unleashing Reasoning and Memory Capabilities in VLA Models for Embodied Visual Tracking},
      author={Jiahang Liu and Yunpeng Qi and Jiazhao Zhang and Minghan Li and Shaoan Wang and Kui Wu and Hanjing Ye and Hong Zhang and Zhibo Chen and Fangwei Zhong and Zhizheng Zhang and He Wang},
      year={2025},
      eprint={2510.07134},
      archivePrefix={arXiv},
      primaryClass={cs.RO},
      url={https://arxiv.org/abs/2510.07134},
}

@inproceedings{sun2025openworld,
  title={Open-World Drone Active Tracking with Goal-Centered Rewards},
  author={Sun, Haowei and Hu, Jinwu and Zhang, Zhirui and Tian, Haoyuan and Xie, Xinze and Wang, Yufeng and Xie, Xiaohua and Lin, Yun and Yu, Zhuliang and Tan, Mingkui},
  booktitle={Advances in Neural Information Processing Systems (NeurIPS)},
  year={2025},
  eprint={2412.00744},
  archivePrefix={arXiv},
  primaryClass={cs.RO},
  url={https://arxiv.org/abs/2412.00744}
}

@article{9521193,
  author={Xi, Mao and Zhou, Yun and Chen, Zheng and Zhou, Wengang and Li, Houqiang},
  journal={IEEE Transactions on Circuits and Systems for Video Technology},
  title={Anti-Distractor Active Object Tracking in {3D} Environments},
  year={2022},
  volume={32},
  number={6},
  pages={3697--3707},
  doi={10.1109/TCSVT.2021.3107153},
  url={https://doi.org/10.1109/TCSVT.2021.3107153}
}

@article{dionigi2024d,
  title={{D-VAT}: End-to-End Visual Active Tracking for Micro Aerial Vehicles},
  author={Dionigi, Alberto and Felicioni, Simone and Leomanni, Mirko and Costante, Gabriele},
  journal={IEEE Robotics and Automation Letters},
  volume={9},
  number={6},
  pages={5046--5053},
  year={2024},
  publisher={IEEE},
  doi={10.1109/LRA.2024.3385700},
  eprint={2308.16874},
  archivePrefix={arXiv},
  primaryClass={cs.RO},
  url={https://arxiv.org/abs/2308.16874}
}

@article{8642452,
  author={Luo, Wenhan and Sun, Peng and Zhong, Fangwei and Liu, Wei and Zhang, Tong and Wang, Yizhou},
  journal={IEEE Transactions on Pattern Analysis and Machine Intelligence},
  title={End-to-End Active Object Tracking and Its Real-World Deployment via Reinforcement Learning},
  year={2020},
  volume={42},
  number={6},
  pages={1317--1332},
  doi={10.1109/TPAMI.2019.2899570},
  url={https://doi.org/10.1109/TPAMI.2019.2899570}
}

@inproceedings{bajcsy2024learning,
  title={Learning Vision-Based Pursuit-Evasion Robot Policies},
  author={Bajcsy, Andrea and Loquercio, Antonio and Kumar, Ashish and Malik, Jitendra},
  booktitle={2024 IEEE International Conference on Robotics and Automation (ICRA)},
  pages={9197--9204},
  year={2024},
  organization={IEEE},
  doi={10.1109/ICRA57147.2024.10610881},
  eprint={2308.16185},
  archivePrefix={arXiv},
  primaryClass={cs.RO},
  url={https://arxiv.org/abs/2308.16185}
}

@inproceedings{zhong2023rspt,
  title={{RSPT}: Reconstruct Surroundings and Predict Trajectory for Generalizable Active Object Tracking},
  author={Zhong, Fangwei and Bi, Xiao and Zhang, Yudi and Zhang, Wei and Wang, Yizhou},
  booktitle={Proceedings of the AAAI Conference on Artificial Intelligence (AAAI)},
  volume={37},
  pages={3705--3714},
  year={2023},
  doi={10.1609/aaai.v37i3.25482},
  eprint={2304.03623},
  archivePrefix={arXiv},
  primaryClass={cs.RO},
  url={https://ojs.aaai.org/index.php/AAAI/article/view/25482}
}

@InProceedings{pmlr-v305-wang25f,
  title = 	 {TrackVLA: Embodied Visual Tracking in the Wild},
  author =       {Wang, Shaoan and Zhang, Jiazhao and Li, Minghan and Liu, Jiahang and Li, Anqi and Wu, Kui and Zhong, Fangwei and Yu, Junzhi and Zhang, Zhizheng and Wang, He},
  booktitle = 	 {Proceedings of The 9th Conference on Robot Learning},
  pages = 	 {4139--4164},
  year = 	 {2025},
  editor = 	 {Lim, Joseph and Song, Shuran and Park, Hae-Won},
  volume = 	 {305},
  series = 	 {Proceedings of Machine Learning Research},
  month = 	 {27--30 Sep},
  publisher =    {PMLR},
  url = 	 {https://proceedings.mlr.press/v305/wang25f.html}
}

@Article{Wu2025,
author={Wu, Qihui
and Li, Jiahao
and Zhou, Fuhui
and Ji, Jiahuan
and Wang, Haoyang
and Liang, Hongtao
and Ma, Kai-Kuang},
title={Cognitive embodied learning for anomaly active target tracking},
journal={Communications Engineering},
year={2025},
month={Nov},
day={27},
volume={4},
number={1},
pages={224},
issn={2731-3395},
doi={10.1038/s44172-025-00556-6},
url={https://doi.org/10.1038/s44172-025-00556-6}
}

@inproceedings{Zhong_2025_ICCV,
    author    = {Zhong, Fangwei and Wu, Kui and Wang, Churan and Chen, Hao and Ci, Hai and Li, Zhoujun and Wang, Yizhou},
    title     = {{UnrealZoo}: Enriching Photo-Realistic Virtual Worlds for Embodied AI},
    booktitle = {Proceedings of the IEEE/CVF International Conference on Computer Vision (ICCV)},
    year      = {2025},
    pages     = {5769--5779},
    eprint={2412.20977},
    archivePrefix={arXiv},
    primaryClass={cs.CV},
    url={https://arxiv.org/abs/2412.20977}
}

@article{sun2023uav,
  title={{UAV}-Ground Visual Tracking: A Unified Dataset and Collaborative Learning Approach},
  author={Sun, Dengdi and Cheng, Leilei and Chen, Song and Li, Chenglong and Xiao, Yun and Luo, Bin},
  journal={IEEE Transactions on Circuits and Systems for Video Technology},
  volume={34},
  number={5},
  pages={3619--3632},
  year={2024},
  publisher={IEEE},
  doi={10.1109/TCSVT.2023.3316990},
  url={https://doi.org/10.1109/TCSVT.2023.3316990}
}

@inproceedings{Li_2020_CVPR,
  author = {Li, Yiming and Fu, Changhong and Ding, Fangqiang and Huang, Ziyuan and Lu, Geng},
  title = {{AutoTrack}: Towards High-Performance Visual Tracking for {UAV} with Automatic Spatio-Temporal Regularization},
  booktitle = {Proceedings of the IEEE/CVF Conference on Computer Vision and Pattern Recognition (CVPR)},
  year = {2020},
  eprint={2003.12949},
  archivePrefix={arXiv},
  primaryClass={cs.CV},
  url={https://arxiv.org/abs/2003.12949}
}

@article{xue2024handling,
  title={Handling Occlusion in {UAV} Visual Tracking with Query-Guided Redetection},
  author={Xue, Yuanliang and Shen, Tao and Jin, Guodong and Tan, Lining and Wang, Nian and Wang, Lianfeng and Gao, Jing},
  journal={IEEE Transactions on Instrumentation and Measurement},
  volume={73},
  pages={1--17},
  year={2024},
  publisher={IEEE},
  doi={10.1109/TIM.2024.3373086},
  url={https://doi.org/10.1109/TIM.2024.3373086}
}

@article{wu2025learning,
  title={Learning an Adaptive and View-Invariant Vision Transformer for Real-Time {UAV} Tracking},
  author={Wu, You and Li, Yongxin and Liu, Mengyuan and Wang, Xucheng and Yang, Xiangyang and Ye, Hengzhou and Zeng, Dan and Zhao, Qijun and Li, Shuiwang},
  journal={IEEE Transactions on Circuits and Systems for Video Technology},
  year={2026},
  volume={36},
  number={2},
  pages={2403--2418},
  publisher={IEEE},
  doi={10.1109/TCSVT.2025.3599856},
  eprint={2412.20002},
  archivePrefix={arXiv},
  primaryClass={cs.CV},
  url={https://arxiv.org/abs/2412.20002}
}

@inproceedings{li2023adaptive,
  title={Adaptive and Background-Aware Vision Transformer for Real-Time {UAV} Tracking},
  author={Li, Shuiwang and Yang, Yangxiang and Zeng, Dan and Wang, Xucheng},
  booktitle={Proceedings of the IEEE/CVF International Conference on Computer Vision (ICCV)},
  pages={13989--14000},
  year={2023},
  doi={10.1109/ICCV51070.2023.01285},
  url={https://doi.org/10.1109/ICCV51070.2023.01285}
}

@article{sun2025refdrone,
  title={RefDrone: A Challenging Benchmark for Referring Expression Comprehension in Drone Scenes},
  author={Sun, Zhichao and Liu, Yepeng and Su, Zhiling and Zhu, Huachao and Gu, Yuliang and Zou, Yuda and Liu, Zelong and Xia, Gui-Song and Du, Bo and Xu, Yongchao},
  journal={arXiv preprint arXiv:2502.00392},
  year={2025},
  eprint={2502.00392},
  archivePrefix={arXiv},
  primaryClass={cs.CV},
  url={https://arxiv.org/abs/2502.00392}
}

@inproceedings{Xue_2025_CVPR,
  author    = {Xue, Chaocan and Zhong, Bineng and Liang, Qihua and Zheng, Yaozong and Li, Ning and Xue, Yuanliang and Song, Shuxiang},
  title     = {Similarity-Guided Layer-Adaptive Vision Transformer for {UAV} Tracking},
  booktitle = {Proceedings of the IEEE/CVF Conference on Computer Vision and Pattern Recognition (CVPR)},
  year      = {2025},
  pages     = {6730--6740},
  eprint={2503.06625},
  archivePrefix={arXiv},
  primaryClass={cs.CV},
  url={https://arxiv.org/abs/2503.06625}
}

@misc{cheang2025gr3technicalreport,
      title={GR-3 Technical Report},
      author={Chilam Cheang and Sijin Chen and Zhongren Cui and Yingdong Hu and Liqun Huang and Tao Kong and Hang Li and Yifeng Li and Yuxiao Liu and Xiao Ma and Hao Niu and Wenxuan Ou and Wanli Peng and Zeyu Ren and Haixin Shi and Jiawen Tian and Hongtao Wu and Xin Xiao and Yuyang Xiao and Jiafeng Xu and Yichu Yang},
      year={2025},
      eprint={2507.15493},
      archivePrefix={arXiv},
      primaryClass={cs.RO},
      url={https://arxiv.org/abs/2507.15493},
}

@misc{octomodelteam2024octoopensourcegeneralistrobot,
      title={Octo: An Open-Source Generalist Robot Policy},
      author={Octo Model Team and Dibya Ghosh and Homer Walke and Karl Pertsch and Kevin Black and Oier Mees and Sudeep Dasari and Joey Hejna and Tobias Kreiman and Charles Xu and Jianlan Luo and You Liang Tan and Lawrence Yunliang Chen and Pannag Sanketi and Quan Vuong and Ted Xiao and Dorsa Sadigh and Chelsea Finn and Sergey Levine},
      year={2024},
      eprint={2405.12213},
      archivePrefix={arXiv},
      primaryClass={cs.RO},
      url={https://arxiv.org/abs/2405.12213},
}

@misc{kim2024openvlaopensourcevisionlanguageactionmodel,
      title={OpenVLA: An Open-Source Vision-Language-Action Model},
      author={Moo Jin Kim and Karl Pertsch and Siddharth Karamcheti and Ted Xiao and Ashwin Balakrishna and Suraj Nair and Rafael Rafailov and Ethan Foster and Grace Lam and Pannag Sanketi and Quan Vuong and Thomas Kollar and Benjamin Burchfiel and Russ Tedrake and Dorsa Sadigh and Sergey Levine and Percy Liang and Chelsea Finn},
      year={2024},
      eprint={2406.09246},
      archivePrefix={arXiv},
      primaryClass={cs.RO},
      url={https://arxiv.org/abs/2406.09246},
}

@misc{intelligence2025pi05visionlanguageactionmodelopenworld,
      title={$\pi_{0.5}$: a Vision-Language-Action Model with Open-World Generalization},
      author={Physical Intelligence and Kevin Black and Noah Brown and James Darpinian and Karan Dhabalia and Danny Driess and Adnan Esmail and Michael Equi and Chelsea Finn and Niccolo Fusai and Manuel Y. Galliker and Dibya Ghosh and Lachy Groom and Karol Hausman and Brian Ichter and Szymon Jakubczak and Tim Jones and Liyiming Ke and Devin LeBlanc and Sergey Levine and Adrian Li-Bell and Mohith Mothukuri and Suraj Nair and Karl Pertsch and Allen Z. Ren and Lucy Xiaoyang Shi and Laura Smith and Jost Tobias Springenberg and Kyle Stachowicz and James Tanner and Quan Vuong and Homer Walke and Anna Walling and Haohuan Wang and Lili Yu and Ury Zhilinsky},
      year={2025},
      eprint={2504.16054},
      archivePrefix={arXiv},
      primaryClass={cs.LG},
      url={https://arxiv.org/abs/2504.16054},
}

@misc{intelligence2025pi06vlalearnsexperience,
      title={$\pi^{*}_{0.6}$: a VLA That Learns From Experience},
      author={Physical Intelligence and Ali Amin and Raichelle Aniceto and Ashwin Balakrishna and Kevin Black and Ken Conley and Grace Connors and James Darpinian and Karan Dhabalia and Jared DiCarlo and Danny Driess and Michael Equi and Adnan Esmail and Yunhao Fang and Chelsea Finn and Catherine Glossop and Thomas Godden and Ivan Goryachev and Lachy Groom and Hunter Hancock and Karol Hausman and Gashon Hussein and Brian Ichter and Szymon Jakubczak and Rowan Jen and Tim Jones and Ben Katz and Liyiming Ke and Chandra Kuchi and Marinda Lamb and Devin LeBlanc and Sergey Levine and Adrian Li-Bell and Yao Lu and Vishnu Mano and Mohith Mothukuri and Suraj Nair and Karl Pertsch and Allen Z. Ren and Charvi Sharma and Lucy Xiaoyang Shi and Laura Smith and Jost Tobias Springenberg and Kyle Stachowicz and Will Stoeckle and Alex Swerdlow and James Tanner and Marcel Torne and Quan Vuong and Anna Walling and Haohuan Wang and Blake Williams and Sukwon Yoo and Lili Yu and Ury Zhilinsky and Zhiyuan Zhou},
      year={2025},
      eprint={2511.14759},
      archivePrefix={arXiv},
      primaryClass={cs.LG},
      url={https://arxiv.org/abs/2511.14759},
}

@misc{black2026pi0visionlanguageactionflowmodel,
      title={$\pi_0$: A Vision-Language-Action Flow Model for General Robot Control},
      author={Kevin Black and Noah Brown and Danny Driess and Adnan Esmail and Michael Equi and Chelsea Finn and Niccolo Fusai and Lachy Groom and Karol Hausman and Brian Ichter and Szymon Jakubczak and Tim Jones and Liyiming Ke and Sergey Levine and Adrian Li-Bell and Mohith Mothukuri and Suraj Nair and Karl Pertsch and Lucy Xiaoyang Shi and James Tanner and Quan Vuong and Anna Walling and Haohuan Wang and Ury Zhilinsky},
      year={2024},
      eprint={2410.24164},
      archivePrefix={arXiv},
      primaryClass={cs.LG},
      url={https://arxiv.org/abs/2410.24164},
}

@misc{brohan2023rt1roboticstransformerrealworld,
      title={RT-1: Robotics Transformer for Real-World Control at Scale},
      author={Anthony Brohan and Noah Brown and Justice Carbajal and Yevgen Chebotar and Joseph Dabis and Chelsea Finn and Keerthana Gopalakrishnan and Karol Hausman and Alex Herzog and Jasmine Hsu and Julian Ibarz and Brian Ichter and Alex Irpan and Tomas Jackson and Sally Jesmonth and Nikhil J Joshi and Ryan Julian and Dmitry Kalashnikov and Yuheng Kuang and Isabel Leal and Kuang-Huei Lee and Sergey Levine and Yao Lu and Utsav Malla and Deeksha Manjunath and Igor Mordatch and Ofir Nachum and Carolina Parada and Jodilyn Peralta and Emily Perez and Karl Pertsch and Jornell Quiambao and Kanishka Rao and Michael Ryoo and Grecia Salazar and Pannag Sanketi and Kevin Sayed and Jaspiar Singh and Sumedh Sontakke and Austin Stone and Clayton Tan and Huong Tran and Vincent Vanhoucke and Steve Vega and Quan Vuong and Fei Xia and Ted Xiao and Peng Xu and Sichun Xu and Tianhe Yu and Brianna Zitkovich},
      year={2023},
      eprint={2212.06817},
      archivePrefix={arXiv},
      primaryClass={cs.RO},
      url={https://arxiv.org/abs/2212.06817},
}

@misc{brohan2023rt2visionlanguageactionmodelstransfer,
      title={RT-2: Vision-Language-Action Models Transfer Web Knowledge to Robotic Control},
      author={Anthony Brohan and Noah Brown and Justice Carbajal and Yevgen Chebotar and Xi Chen and Krzysztof Choromanski and Tianli Ding and Danny Driess and Avinava Dubey and Chelsea Finn and Pete Florence and Chuyuan Fu and Montse Gonzalez Arenas and Keerthana Gopalakrishnan and Kehang Han and Karol Hausman and Alexander Herzog and Jasmine Hsu and Brian Ichter and Alex Irpan and Nikhil Joshi and Ryan Julian and Dmitry Kalashnikov and Yuheng Kuang and Isabel Leal and Lisa Lee and Tsang-Wei Edward Lee and Sergey Levine and Yao Lu and Henryk Michalewski and Igor Mordatch and Karl Pertsch and Kanishka Rao and Krista Reymann and Michael Ryoo and Grecia Salazar and Pannag Sanketi and Pierre Sermanet and Jaspiar Singh and Anikait Singh and Radu Soricut and Huong Tran and Vincent Vanhoucke and Quan Vuong and Ayzaan Wahid and Stefan Welker and Paul Wohlhart and Jialin Wu and Fei Xia and Ted Xiao and Peng Xu and Sichun Xu and Tianhe Yu and Brianna Zitkovich},
      year={2023},
      eprint={2307.15818},
      archivePrefix={arXiv},
      primaryClass={cs.RO},
      url={https://arxiv.org/abs/2307.15818},
}

@misc{li2025grrlgoingdexterousprecise,
      title={GR-RL: Going Dexterous and Precise for Long-Horizon Robotic Manipulation},
      author={Yunfei Li and Xiao Ma and Jiafeng Xu and Yu Cui and Zhongren Cui and Zhigang Han and Liqun Huang and Tao Kong and Yuxiao Liu and Hao Niu and Wanli Peng and Jingchao Qiao and Zeyu Ren and Haixin Shi and Zhi Su and Jiawen Tian and Yuyang Xiao and Shenyu Zhang and Liwei Zheng and Hang Li and Yonghui Wu},
      year={2025},
      eprint={2512.01801},
      archivePrefix={arXiv},
      primaryClass={cs.RO},
      url={https://arxiv.org/abs/2512.01801},
}

@misc{lykov2025cognitivedronevlamodelevaluation,
      title={CognitiveDrone: A VLA Model and Evaluation Benchmark for Real-Time Cognitive Task Solving and Reasoning in UAVs},
      author={Artem Lykov and Valerii Serpiva and Muhammad Haris Khan and Oleg Sautenkov and Artyom Myshlyaev and Grik Tadevosyan and Yasheerah Yaqoot and Dzmitry Tsetserukou},
      year={2025},
      eprint={2503.01378},
      archivePrefix={arXiv},
      primaryClass={cs.RO},
      url={https://arxiv.org/abs/2503.01378},
}

@inproceedings{11149880,
  author={Li, JinHai and Chen, Peng and Li, MoHan and Ren, LuYi},
  booktitle={2025 40th Youth Academic Annual Conference of Chinese Association of Automation (YAC)},
  title={{IB-AMG}: Aircraft Mission Generation With Inference-based Vision-Language-Action Model},
  year={2025},
  pages={2323--2328},
  doi={10.1109/YAC66630.2025.11149880},
  url={https://doi.org/10.1109/YAC66630.2025.11149880}
}

@misc{liu2025indooruavbenchmarkingvisionlanguageuav,
      title={IndoorUAV: Benchmarking Vision-Language UAV Navigation in Continuous Indoor Environments},
      author={Xu Liu and Yu Liu and Hanshuo Qiu and Yang Qirong and Zhouhui Lian},
      year={2025},
      eprint={2512.19024},
      archivePrefix={arXiv},
      primaryClass={cs.RO},
      url={https://arxiv.org/abs/2512.19024},
}

@misc{serpiva2025racevlavlabasedracingdrone,
      title={RaceVLA: VLA-based Racing Drone Navigation with Human-like Behaviour},
      author={Valerii Serpiva and Artem Lykov and Artyom Myshlyaev and Muhammad Haris Khan and Ali Alridha Abdulkarim and Oleg Sautenkov and Dzmitry Tsetserukou},
      year={2025},
      eprint={2503.02572},
      archivePrefix={arXiv},
      primaryClass={cs.RO},
      url={https://arxiv.org/abs/2503.02572},
}

@misc{wang2025uavflowcolosseorealworldbenchmark,
      title={UAV-Flow Colosseo: A Real-World Benchmark for Flying-on-a-Word UAV Imitation Learning},
      author={Xiangyu Wang and Donglin Yang and Yue Liao and Wenhao Zheng and wenjun wu and Bin Dai and Hongsheng Li and Si Liu},
      year={2025},
      eprint={2505.15725},
      archivePrefix={arXiv},
      primaryClass={cs.RO},
      url={https://arxiv.org/abs/2505.15725},
}

@inproceedings{10974117,
  author={Sautenkov, Oleg and Yaqoot, Yasheerah and Lykov, Artem and Mustafa, Muhammad Ahsan and Tadevosyan, Grik and Akhmetkazy, Aibek and Cabrera, Miguel Altamirano and Martynov, Mikhail and Karaf, Sausar and Tsetserukou, Dzmitry},
  booktitle={2025 20th ACM/IEEE International Conference on Human-Robot Interaction (HRI)},
  title={{UAV-VLA}: Vision-Language-Action System for Large Scale Aerial Mission Generation},
  year={2025},
  pages={1588--1592},
  doi={10.1109/HRI61500.2025.10974117},
  url={https://doi.org/10.1109/HRI61500.2025.10974117}
}

@misc{wu2025vlaanefficientonboardvisionlanguageaction,
      title={VLA-AN: An Efficient and Onboard Vision-Language-Action Framework for Aerial Navigation in Complex Environments},
      author={Yuze Wu and Mo Zhu and Xingxing Li and Yuheng Du and Yuxin Fan and Wenjun Li and Zhichao Han and Xin Zhou and Fei Gao},
      year={2025},
      eprint={2512.15258},
      archivePrefix={arXiv},
      primaryClass={cs.RO},
      url={https://arxiv.org/abs/2512.15258},
}

@misc{sun2026airvlavisionlanguageactionsystemsaerial,
      title={AIR-VLA: Vision-Language-Action Systems for Aerial Manipulation},
      author={Jianli Sun and Bin Tian and Qiyao Zhang and Chengxiang Li and Zihan Song and Zhiyong Cui and Yisheng Lv and Yonglin Tian},
      year={2026},
      eprint={2601.21602},
      archivePrefix={arXiv},
      primaryClass={cs.RO},
      url={https://arxiv.org/abs/2601.21602},
}

@misc{sun2026autoflyvisionlanguageactionmodeluav,
      title={AutoFly: Vision-Language-Action Model for UAV Autonomous Navigation in the Wild},
      author={Xiaolou Sun and Wufei Si and Wenhui Ni and Yuntian Li and Dongming Wu and Fei Xie and Runwei Guan and He-Yang Xu and Henghui Ding and Yuan Wu and Yutao Yue and Yongming Huang and Hui Xiong},
      year={2026},
      eprint={2602.09657},
      archivePrefix={arXiv},
      primaryClass={cs.RO},
      url={https://arxiv.org/abs/2602.09657},
}

@misc{huang2026navdreamervideomodelszeroshot,
      title={NavDreamer: Video Models as Zero-Shot 3D Navigators},
      author={Xijie Huang and Weiqi Gai and Tianyue Wu and Congyu Wang and Zhiyang Liu and Xin Zhou and Yuze Wu and Fei Gao},
      year={2026},
      eprint={2602.09765},
      archivePrefix={arXiv},
      primaryClass={cs.RO},
      url={https://arxiv.org/abs/2602.09765},
}

@misc{li2024cogactfoundationalvisionlanguageactionmodel,
      title={CogACT: A Foundational Vision-Language-Action Model for Synergizing Cognition and Action in Robotic Manipulation},
      author={Qixiu Li and Yaobo Liang and Zeyu Wang and Lin Luo and Xi Chen and Mozheng Liao and Fangyun Wei and Yu Deng and Sicheng Xu and Yizhong Zhang and Xiaofan Wang and Bei Liu and Jianlong Fu and Jianmin Bao and Dong Chen and Yuanchun Shi and Jiaolong Yang and Baining Guo},
      year={2024},
      eprint={2411.19650},
      archivePrefix={arXiv},
      primaryClass={cs.RO},
      url={https://arxiv.org/abs/2411.19650},
}

@misc{han2024dualprocessvlaefficient,
      title={A Dual Process VLA: Efficient Robotic Manipulation Leveraging VLM},
      author={ByungOk Han and Jaehong Kim and Jinhyeok Jang},
      year={2024},
      eprint={2410.15549},
      archivePrefix={arXiv},
      primaryClass={cs.RO},
      url={https://arxiv.org/abs/2410.15549},
}

@article{song2025hume,
  title={Hume: Introducing System-2 Thinking in Visual-Language-Action Model},
  author={Song, Haoming and Qu, Delin and Yao, Yuanqi and Chen, Qizhi and Lv, Qi and Tang, Yiwen and Shi, Modi and Ren, Guanghui and Yao, Maoqing and Zhao, Bin and others},
  journal={arXiv preprint arXiv:2505.21432},
  year={2025},
  eprint={2505.21432},
  archivePrefix={arXiv},
  primaryClass={cs.RO},
  url={https://arxiv.org/abs/2505.21432}
}

@article{tian2025uavs,
  title={{UAVs} Meet {LLMs}: Overviews and Perspectives Toward Agentic Low-Altitude Mobility},
  author={Tian, Yonglin and Lin, Fei and Li, Yiduo and Zhang, Tengchao and Zhang, Qiyao and Fu, Xuan and Huang, Jun and Dai, Xingyuan and Wang, Yutong and Tian, Chunwei and others},
  journal={Information Fusion},
  volume={122},
  pages={103158},
  year={2025},
  publisher={Elsevier},
  doi={10.1016/j.inffus.2025.103158},
  eprint={2501.02341},
  archivePrefix={arXiv},
  primaryClass={cs.RO},
  url={https://arxiv.org/abs/2501.02341}
}

@article{10445025,
  author={Sun, Nianyi and Zhao, Jin and Shi, Qing and Liu, Chang and Liu, Peng},
  journal={IEEE Transactions on Industrial Informatics},
  title={Moving Target Tracking by Unmanned Aerial Vehicle: A Survey and Taxonomy},
  year={2024},
  volume={20},
  number={5},
  pages={7056--7068},
  doi={10.1109/TII.2024.3363084},
  url={https://doi.org/10.1109/TII.2024.3363084}
}

@article{khawaja2025survey,
  title={A Survey on Detection, Classification, and Tracking of {UAVs} Using Radar and Communications Systems},
  author={Khawaja, Wahab and Ezuma, Martins and Semkin, Vasilii and Erden, Fatih and Ozdemir, Ozgur and Guvenc, Ismail},
  journal={IEEE Communications Surveys \& Tutorials},
  volume={28},
  pages={3272--3310},
  year={2025},
  publisher={IEEE},
  doi={10.1109/COMST.2025.3554613},
  eprint={2402.05909},
  archivePrefix={arXiv},
  primaryClass={cs.IT},
  url={https://arxiv.org/abs/2402.05909}
}

@article{jayaweera2020dynamic,
  title={A Dynamic Artificial Potential Field ({D-APF}) {UAV} Path Planning Technique for Following Ground Moving Targets},
  author={Jayaweera, Herath M. P. C. and Hanoun, Samer},
  journal={IEEE Access},
  volume={8},
  pages={192760--192776},
  year={2020},
  publisher={IEEE},
  doi={10.1109/ACCESS.2020.3032929},
  url={https://doi.org/10.1109/ACCESS.2020.3032929}
}

@misc{zhai2025ignitingvlmsembodiedspace,
      title={Igniting VLMs toward the Embodied Space},
      author={Andy Zhai and Brae Liu and Bruno Fang and Chalse Cai and Ellie Ma and Ethan Yin and Hao Wang and Hugo Zhou and James Wang and Lights Shi and Lucy Liang and Make Wang and Qian Wang and Roy Gan and Ryan Yu and Shalfun Li and Starrick Liu and Sylas Chen and Vincent Chen and Zach Xu},
      year={2025},
      eprint={2509.11766},
      archivePrefix={arXiv},
      primaryClass={cs.RO},
      url={https://arxiv.org/abs/2509.11766},
}

@article{chen2026track,
  title={Track A*: Fast Visibility-Aware Trajectory Planning for Active Target Tracking},
  author={Chen, Hanxuan and Wang, Kangli and Pei, Ji},
  journal={arXiv preprint arXiv:2605.05338},
  year={2026},
  eprint={2605.05338},
  archivePrefix={arXiv},
  primaryClass={cs.RO},
  url={https://arxiv.org/abs/2605.05338}
}

@article{chen2026visionlanguageuavs,
  title={Vision-and-Language Navigation for UAVs: Progress, Challenges, and a Research Roadmap},
  author={Chen, Hanxuan and Zheng, Jie and Yang, Siqi and Zeng, Tianle and Feng, Siwei and Cheng, Songsheng and Ren, Ruilong and Guo, Hanzhong and Yuan, Shuai and Wang, Xiangyue and others},
  journal={arXiv preprint arXiv:2604.13654},
  year={2026},
  eprint={2604.13654},
  archivePrefix={arXiv},
  primaryClass={cs.RO},
  url={https://arxiv.org/abs/2604.13654}
}

@article{chen2026cosfly,
  title={CosFly: Plan in the Matrix, Fly in the World},
  author={Chen, Hanxuan and Wang, Xiangyue and Cheng, Songsheng and Ren, Ruilong and Zheng, Jie and Yuan, Shuai and Zeng, Tianle and Guo, Hanzhong and Li, Binbo and Wang, Kangli and others},
  journal={arXiv preprint arXiv:2605.19120},
  year={2026},
  eprint={2605.19120},
  archivePrefix={arXiv},
  primaryClass={cs.RO},
  url={https://arxiv.org/abs/2605.19120}
}

@article{wang2026cosfly,
  title={CosFly-Track: A Large-Scale Multi-Modal Dataset for UAV Visual Tracking via Multi-Constraint Trajectory Optimization},
  author={Wang, Xiangyue and Chen, Hanxuan and Cheng, Songsheng and Ren, Ruilong and Zheng, Jie and Yuan, Shuai and Zeng, Tianle and Guo, Hanzhong and Wang, Kangli and Pei, Ji},
  journal={arXiv preprint arXiv:2605.17776},
  year={2026},
  eprint={2605.17776},
  archivePrefix={arXiv},
  primaryClass={cs.CV},
  url={https://arxiv.org/abs/2605.17776}
}

@article{zeng2025ezreal,
  title={EZREAL: Enhancing Zero-Shot Outdoor Robot Navigation toward Distant Targets under Varying Visibility},
  author={Zeng, Tianle and Peng, Jianwei and Ye, Hanjing and Chen, Guangcheng and Luo, Senzi and Zhang, Hong},
  journal={arXiv preprint arXiv:2509.13720},
  year={2025},
  eprint={2509.13720},
  archivePrefix={arXiv},
  primaryClass={cs.RO},
  url={https://arxiv.org/abs/2509.13720}
}

@article{zeng2026carla,
  title={CARLA-Air: Fly Drones Inside a CARLA World--A Unified Infrastructure for Air-Ground Embodied Intelligence},
  author={Zeng, Tianle and Wen, Yanci and Zhang, Hong},
  journal={arXiv preprint arXiv:2603.28032},
  year={2026},
  eprint={2603.28032},
  archivePrefix={arXiv},
  primaryClass={cs.RO},
  url={https://arxiv.org/abs/2603.28032}
}

@article{zeng2026can,
  title={Can Aerial VLA Models Cooperate? Evaluating Closed-Loop Air-Ground Coordination with CARLA-Air},
  author={Zeng, Tianle and Wen, Yanci and Yu, Xueang and Zhang, Hong},
  journal={arXiv preprint arXiv:2605.31066},
  year={2026},
  eprint={2605.31066},
  archivePrefix={arXiv},
  primaryClass={cs.RO},
  url={https://arxiv.org/abs/2605.31066}
}

@article{aircop2025,
  title={AirCopBench: A Benchmark for Multi-drone Collaborative Embodied Perception and Reasoning},
  author={Zha, Jirong and Fan, Yuxuan and Zhang, Tianyu and Chen, Geng and Chen, Yingfeng and Gao, Chen and Chen, Xinlei},
  journal={arXiv preprint arXiv:2511.11025},
  year={2025},
  eprint={2511.11025},
  archivePrefix={arXiv},
  primaryClass={cs.CV},
  url={https://arxiv.org/abs/2511.11025}
}

@article{airspatial2025,
  title={AirSpatialBot: A Spatially-Aware Aerial Agent for Fine-Grained Vehicle Attribute Recognition and Retrieval},
  author={Zhou, Yue and Ding, Ran and Yang, Xue and Xue, Jiang and Liu, Xingzhao},
  journal={IEEE Transactions on Geoscience and Remote Sensing},
  year={2025},
  volume={63},
  pages={1--15},
  doi={10.1109/TGRS.2025.3570895},
  eprint={2601.01416},
  archivePrefix={arXiv},
  primaryClass={cs.CV},
  url={https://arxiv.org/abs/2601.01416}
}

@article{avimath2025,
  title={Multimodal Mathematical Reasoning Embedded in Aerial Vehicle Imagery: Benchmarking, Analysis, and Exploration},
  author={Zhou, Yue and Feng, Litong and Lan, Mengcheng and Yang, Xue and Li, Qingyun and Ke, Yiping and Xue, Jiang and Zhang, Wayne},
  journal={ISPRS Journal of Photogrammetry and Remote Sensing},
  year={2025},
  eprint={2509.10059},
  archivePrefix={arXiv},
  primaryClass={cs.CV},
  url={https://arxiv.org/abs/2509.10059}
}

@article{capera2025,
  title={CapERA: Captioning Events in Aerial Videos},
  author={Bashmal, Laila and Bazi, Yakoub and Al Rahhal, Mohamad Mahmoud and Zuair, Mansour and Melgani, Farid},
  journal={Remote Sensing},
  volume={15},
  number={8},
  pages={2139},
  year={2023},
  publisher={MDPI},
  doi={10.3390/rs15082139},
  url={https://researchr.org/publication/BashmalBRZM23}
}

@article{hrvqa2024,
  title={HRVQA: A Visual Question Answering benchmark for high-resolution aerial images},
  author={Li, Kun and Vosselman, George and Yang, Michael Ying},
  journal={ISPRS Journal of Photogrammetry and Remote Sensing},
  volume={214},
  pages={65--81},
  year={2024},
  publisher={Elsevier},
  doi={10.1016/j.isprsjprs.2024.06.002},
  eprint={2301.09460},
  archivePrefix={arXiv},
  primaryClass={cs.CV},
  url={https://arxiv.org/abs/2301.09460}
}

@inproceedings{open3dvqa2025,
  title={Open3D-VQA: A Benchmark for Embodied Spatial Concept Reasoning with Multimodal Large Language Model in Open Space},
  author={Du, Weichao and Zhang, Tianyu and Wang, Lixiang and Gao, Chen and Liu, Yong and Chen, Xinlei and Pang, Yueting},
  booktitle={Proceedings of the 33rd ACM International Conference on Multimedia (ACM MM)},
  year={2025},
  doi={10.1145/3746027.3758219},
  eprint={2503.11094},
  archivePrefix={arXiv},
  primaryClass={cs.CV},
  url={https://arxiv.org/abs/2503.11094}
}

@article{alpamayo2025,
  title={Alpamayo-R1: Bridging Reasoning and Action Prediction for Generalizable Autonomous Driving in the Long Tail},
  author={NVIDIA Alpamayo-R1 Team and Pavone, Marco},
  journal={arXiv preprint arXiv:2511.00088},
  year={2025},
  eprint={2511.00088},
  archivePrefix={arXiv},
  primaryClass={cs.RO},
  url={https://arxiv.org/abs/2511.00088}
}

@article{zhang2026uavtrackvla,
  title={{UAV-Track VLA}: Embodied Aerial Tracking via Vision-Language-Action Models},
  author={Zhang, Qiyao and Zheng, Shuhua and Sun, Jianli and Li, Chengxiang and Wu, Xianke and Song, Zihan and Cui, Zhiyong and Lv, Yisheng and Tian, Yonglin},
  journal={arXiv preprint arXiv:2604.02241},
  year={2026},
  eprint={2604.02241},
  archivePrefix={arXiv},
  primaryClass={cs.CV},
  url={https://arxiv.org/abs/2604.02241}
}

@article{xu2026aerialvla,
  title={{AerialVLA}: A Vision-Language-Action Model for UAV Navigation via Minimalist End-to-End Control},
  author={Xu, Peng and Deng, Zhengnan and Deng, Jiayan and Gu, Zonghua and Wan, Shaohua},
  journal={arXiv preprint arXiv:2603.14363},
  year={2026},
  eprint={2603.14363},
  archivePrefix={arXiv},
  primaryClass={cs.CV},
  url={https://arxiv.org/abs/2603.14363}
}

@article{zhao2026worldvln,
  title={{WorldVLN}: Autoregressive World Action Model for Aerial Vision-Language Navigation},
  author={Zhao, Baining and Xu, Jiacheng and Feng, Weicheng and Zhang, Xin and Wang, Zhaolu and Wang, Haoyang and Ji, Shilong and Wang, Ziyou and Fang, Jianjie and Zheng, Zhiheng and Zhang, Weichen and Shang, Yu and Wu, Wei and Gao, Chen and Chen, Xinlei and Li, Yong},
  journal={arXiv preprint arXiv:2605.15964},
  year={2026},
  eprint={2605.15964},
  archivePrefix={arXiv},
  primaryClass={cs.RO},
  url={https://arxiv.org/abs/2605.15964}
}

@article{guo2026awarevln,
  title={{AwareVLN}: Reasoning with Self-awareness for Vision-Language Navigation},
  author={Guo, Wenxuan and Xu, Xiuwei and Liu, Yichen and Li, Xiangyu and Yin, Hang and Chen, Huangxing and Zheng, Wenzhao and Feng, Jianjiang and Zhou, Jie and Lu, Jiwen},
  journal={arXiv preprint arXiv:2605.22816},
  year={2026},
  eprint={2605.22816},
  archivePrefix={arXiv},
  primaryClass={cs.CV},
  url={https://arxiv.org/abs/2605.22816}
}

@misc{anthropic2026claudeopus46,
  title={{Claude Opus 4.6}},
  author={{Anthropic}},
  year={2026},
  month={February},
  url={https://www.anthropic.com/research/claude-opus-4-6}
}

@misc{googledeepmind2026gemini31pro,
  title={{Gemini 3.1 Pro Model Card}},
  author={{Google DeepMind}},
  year={2026},
  month={February},
  url={https://deepmind.google/models/model-cards/gemini-3-1-pro/}
}

@article{qwen2025qwen3vl,
  title={{Qwen3-VL Technical Report}},
  author={{Qwen Team}},
  journal={arXiv preprint arXiv:2511.21631},
  year={2025},
  eprint={2511.21631},
  archivePrefix={arXiv},
  primaryClass={cs.CV},
  url={https://arxiv.org/abs/2511.21631}
}

@misc{qwen2026qwen35,
  title={{Qwen3.5}: Towards Native Multimodal Agents},
  author={{Qwen Team}},
  year={2026},
  month={February},
  url={https://qwen.ai/blog?id=qwen3.5}
}

@article{schulman2017proximal,
  title={Proximal Policy Optimization Algorithms},
  author={Schulman, John and Wolski, Filip and Dhariwal, Prafulla and Radford, Alec and Klimov, Oleg},
  journal={arXiv preprint arXiv:1707.06347},
  year={2017},
  eprint={1707.06347},
  archivePrefix={arXiv},
  primaryClass={cs.LG},
  url={https://arxiv.org/abs/1707.06347}
}

@article{shao2024deepseekmath,
  title={{DeepSeekMath}: Pushing the Limits of Mathematical Reasoning in Open Language Models},
  author={Shao, Zhihong and Wang, Peiyi and Zhu, Qihao and Xu, Runxin and Song, Junxiao and Bi, Xiao and Zhang, Haowei and Zhang, Mingchuan and Li, Y. K. and Wu, Y. and Guo, Daya},
  journal={arXiv preprint arXiv:2402.03300},
  year={2024},
  eprint={2402.03300},
  archivePrefix={arXiv},
  primaryClass={cs.CL},
  url={https://arxiv.org/abs/2402.03300}
}

@inproceedings{lipman2023flowmatching,
  title={Flow Matching for Generative Modeling},
  author={Lipman, Yaron and Chen, Ricky T. Q. and Ben-Hamu, Heli and Nickel, Maximilian and Le, Matt},
  booktitle={International Conference on Learning Representations (ICLR)},
  year={2023},
  eprint={2210.02747},
  archivePrefix={arXiv},
  primaryClass={cs.LG},
  url={https://arxiv.org/abs/2210.02747}
}

@inproceedings{chi2023diffusionpolicy,
  title={Diffusion Policy: Visuomotor Policy Learning via Action Diffusion},
  author={Chi, Cheng and Feng, Siyuan and Du, Yilun and Xu, Zhenjia and Cousineau, Eric and Burchfiel, Benjamin and Song, Shuran},
  booktitle={Proceedings of Robotics: Science and Systems (RSS)},
  year={2023},
  doi={10.15607/RSS.2023.XIX.026},
  eprint={2303.04137},
  archivePrefix={arXiv},
  primaryClass={cs.RO},
  url={https://arxiv.org/abs/2303.04137}
}

@inproceedings{hu2022lora,
  title={{LoRA}: Low-Rank Adaptation of Large Language Models},
  author={Hu, Edward J. and Shen, Yelong and Wallis, Phillip and Allen-Zhu, Zeyuan and Li, Yuanzhi and Wang, Shean and Wang, Lu and Chen, Weizhu},
  booktitle={International Conference on Learning Representations (ICLR)},
  year={2022},
  eprint={2106.09685},
  archivePrefix={arXiv},
  primaryClass={cs.CL},
  url={https://arxiv.org/abs/2106.09685}
}

@inproceedings{peebles2023dit,
  title={Scalable Diffusion Models with Transformers},
  author={Peebles, William and Xie, Saining},
  booktitle={Proceedings of the IEEE/CVF International Conference on Computer Vision (ICCV)},
  pages={4195--4205},
  year={2023},
  doi={10.1109/ICCV51070.2023.00387},
  eprint={2212.09748},
  archivePrefix={arXiv},
  primaryClass={cs.CV},
  url={https://arxiv.org/abs/2212.09748}
}
